\RequirePackage{fix-cm}
\documentclass[twocolumn]{svjour3}          
\smartqed  
\usepackage{url}
\usepackage{color}
\usepackage[font=footnotesize]{subfig}
\usepackage{graphicx}                                                        
\usepackage{float} 
\usepackage{multirow}
\usepackage{subfloat}
\usepackage{amsmath}
\usepackage{amssymb}
\usepackage{tabularx}
\usepackage{booktabs}
\usepackage{hyperref}
\definecolor{rewriting_color}{RGB}{120, 90, 15}
\usepackage{hyperref}
\hypersetup{
	colorlinks=true,
	linkcolor=blue,
	filecolor=blue,      
	urlcolor=red,
	citecolor=green,
}
\usepackage[T1]{fontenc}
\usepackage{babel}
\usepackage[abs]{overpic}
\usepackage{rotating}

\begin{document}
\begin{sloppypar}

\title{You Can Mask More For Extremely Low-Bitrate Image Compression}
\author{Anqi Li$^{1,2\dag}$ \and Feng Li$^{3\dag}$ \and Jiaxin Han$^{1,2}$ \and Huihui Bai$^{1,2}$* \and Runmin~Cong$^{4}$ \and Chunjie Zhang$^{1,2}$ \and Meng Wang$^{3}$ \and Weisi Lin$^{5}$ \and Yao Zhao$^{1,2}$}
\institute{ Anqi Li:  
            lianqi@bjtu.edu.cn\at
            Feng Li:  
            fengli@hfut.edu.cn\at
            Jiaxin Han:  
            hanjiaxin@bjtu.edu.cn\at
            Huihui Bai: 
            hhbai@bjtu.edu.cn\at
            Runmin Cong: 
            rmcong@sdu.edu.cn\at
            Chunjie Zhang: 
            ivazhangchunjie@gmail.com\at
            Meng Wang: 
            eric.mengwang@gmail.com\at
            Weisi Lin: 
            wslin@ntu.edu.sg\at
            Yao Zhao: 
            yzhao@bjtu.edu.cn \at
             $^{1}$ Institute of Information Science, Beijing Jiaotong University \at
            $^{2}$ Beijing Key Laboratory of Advanced Information Science and Network Technology \at
            $^{3}$ School of Computer Science and Engineering, Hefei University of Technology \at
            $^{4}$ School of Control Science and Engineering, Shandong University \at
            $^{5}$ School of Computer Science and Engineering, Nanyang Technological University \at
            $\dag$~Equal contribution.\at
            * Corresponding author.}
\date{Received: date / Accepted: date}

\maketitle

\begin{abstract}
Learned image compression (LIC) methods have experienced significant progress during recent years. However, these methods are primarily dedicated to optimizing the rate-distortion (R-D) performance~at~medium~and~high~bitrates ($>$~0.1~bits per~pixel (bpp)), while research on extremely low bitrates is limited. 
Besides, existing methods fail to explicitly explore the image structure and texture components crucial for image compression, treating them equally alongside uninformative components in networks. 
This can cause severe perceptual quality degradation, especially under low-bitrate scenarios. 
In this work, inspired by the success of pre-trained masked autoencoders (MAE) in many downstream tasks, 
we propose to rethink its mask sampling strategy from structure and texture perspectives for high redundancy reduction and discriminative feature representation, further unleashing the potential of LIC methods. Therefore, we present a dual-adaptive masking approach (DA-Mask) that samples visible patches based on the structure and texture distributions of original images. We combine DA-Mask and pre-trained MAE in masked image modeling (MIM) as an initial compressor that abstracts informative semantic context and texture representations. Such a pipeline can well cooperate with LIC networks to achieve further secondary compression while preserving promising reconstruction quality. 
Consequently, we propose a simple yet effective masked compression model (MCM), the first framework that unifies MIM and LIC end-to-end for extremely low-bitrate image compression. Extensive experiments have demonstrated that our approach outperforms recent state-of-the-art methods in R-D performance, visual quality, and downstream applications, at very low bitrates. Our code is available at \href{https://github.com/lianqi1008/MCM.git}{https://github.com/lianqi1008/MCM.git}.

\end{abstract}

\section{Introduction}
\label{intro}
Image compression is a fundamental and crucial technique in image processing, with the goal of signal redundancy removal, to pursue efficient storage and transmission while maintaining high perceptual quality, which is ubiquitous in real-world applications. Classical compression standards such as JPEG~\cite{wallace1990overview}, JPEG2000~\cite{taubman2002jpeg2000}, BPG~\cite{BPG}, VVC~\cite{bross2021developments} adopt the sequential paradigm of ``transformation, quantization,~and entropy~coding'', where each is separately optimized with hand-crafted rules, making it hard to be adapted to diverse image content. 

During recent years, deep learning-based approaches~\cite{balle2017endtoend,balle2018variational,minnen2018joint,minnen2020channel,zhu2022unified} have dominated the image compression field due to their end-to-end optimization. Typical learned image compression (LIC) methods follow the variational autoencoder (VAE) design in~\cite{balle2018variational} and construct different autoencoder architectures~\cite{Cai2018LearningAS,jiang2017end,li2018learning,cheng2019learning} or entropy models~\cite{minnen2018joint,lee2018contextadaptive,Yichen_2022_ICLR} to optimize for rate-distortion (R-D) performance measured by the Peak Signal-to-Noise Ratio (PSNR) and Multi-Scale-Structural Similarity Index Measure (MS-SSIM). These methods prioritize improving such metric scores with more competitive bitrates as traditional codecs (\emph{i.e.} BPG and VVC), while the compression at very low bitrates ($\leq$ 0.1 bpp) still lacks sufficient research attention.
Agustsson~\emph{et al.}~\cite{GANcompression} present a generative adversarial network (GAN)-based compression framework that achieves dramatic bitrate savings. The results tend to maintain the high-level semantics but with significant deviated details of original inputs. Besides, the structure and texture components of an image have different tolerance for compression distortion, especially at low bitrates. Nevertheless, both traditional and most existing LIC methods consider all image components equally during their compression process. Due to the lack of explicit exploration of such two components, directly applying these methods to extremely low-bitrate compression scenarios can cause serious degradation of reconstruction quality. 
Several methods~\cite{chang2022conceptual,chang2022consistency} introduce conceptual compression which characterizes visual structure and texture into deep compact representations for individual compression but show limited perceptual quality at low bitrates. How to adapt both two components to achieve extremely low-bitrate compression while preserving high-quality reconstruction is still a serious challenge.

Recently, masked autoencoder (MAE)~\cite{MAE} has exhibited great success on various computer vision tasks. Its superiority can be attributed to the masked sampling and self-supervised pre-training framework, which masks random patches from the input image and only operates on the remaining visible patches to reconstruct masked ones. MAE demonstrates that masking a high portion of patches can effectively reduce data redundancy, thus contributing to efficient learning. Therefore, it is natural for us to consider whether can combine MAE and LIC to achieve extremely low-bitrate compression. However, simply implementing such random masking in image compression is sub-optimal, as it may discard some important texture or structure information and instead preserve some meaningless patches, resulting in less discriminative representation and inaccurate reconstruction.

Based on the above analysis, in this paper, we propose the masked compression model (MCM) which borrows the idea of MAE to unleash
the potential of LIC at low bitrate conditions. Our MCM stems from the fact that the appearance of visual objects is depicted through structures and textures, and the complexity of such two components heavily affects visual analysis and content synthesis. Therefore, we first present a dual-adaptive masking (DA-Mask) approach, which performs patch sampling based on structure and texture distributions. To be specific, we simultaneously quantify the edge information and semantic context to model the texture and structure complexity of all patches in an image. Then, we adopt probability sampling to select visible patches by estimating the categorical distribution over all patches conditioned on both complexity scores. In this way, compared to the random sampling strategy in MAE, our DA-Mask can adaptively sample more informative patches that are conducive to image reconstruction while achieving effective redundancy removal. Moreover, by incorporating the advantages of DA-Mask and pre-trained MAE into masked image modeling (MIM), we enable an initial compressor to abstract meaningful semantic context and texture representation. Subsequently, operating on the learned representations, a simple yet effective deep codec is utilized to conduct further secondary compression and high-quality image reconstruction. Therefore, our MCM can be seen as an end-to-end two-stage framework that unifies MIM and LIC for extremely low-bitrate compression. Extensive experimental results demonstrate the effectiveness of our method which outperforms the existing state-of-the-art methods in perceptual quality at very low bitrates. 
We also combine MCM with existing segmentation and detection backbones to evaluate the performance of MCM on downstream tasks and our MCM achieves better results than various LIC methods. 
The contributions of this paper are summarized as follows.
\begin{itemize}
    \item We propose a novel masked compression model (MCM) for extremely low-bitrate compression. To the best of our knowledge, this is the first framework that generalizes the self-supervised pre-trained MAE to LIC.

    \item We propose a dual-adaptive masking (DA-Mask) approach based on structure and texture distributions, which can adaptively sample more informative patches while achieving effective redundancy removal.

    \item We combine DA-Mask and pre-trained MAE into masked image modeling as an initial compressor to abstract meaningful semantic context and texture representations. Such a pipeline can well cooperate with LIC networks to achieve two-step compression. 

    \item Extensive experiments demonstrate the superiority of the proposed method against recent state-of-the-art LIC methods in visual quality at very low bitrates. MCM also outperforms previous methods on downstream tasks.
\end{itemize}

\section{Related Work}
\subsection{Learned Image Compression}
Since Ball{\'e}~\emph{et al.}~\cite{ball2017endtoend} first propose the pioneering end-to-end image compression method using convolutional neural network (CNN), deep learning models have been widely used in image compression and demonstrated remarkable achievement. \cite{balle2018variational} introduces a scale hyperprior into~\cite{ball2017endtoend} to capture
spatial dependencies in the latent representation, which provides a general variational autoencoder (VAE) compression pipeline, inspiring later research. Minner~\emph{et al.}~\cite{minnen2018joint} combine an autoregressive model with the hyperprior in a joint framework to predict latents from their causal context. 
Lee~\emph{et al.}~\cite{lee2018contextadaptive} exploit bit-consuming and bit-free contexts to build a context-adaptive entropy model that more accurately estimates the distribution of each latent representation. Cheng~\emph{et al.}~\cite{cheng2019learning} 
leverage discretized Gaussian mixture likelihoods-based entropy model and attention modules to enhance the R-D performance. Apart from these CNN-based models, motivated by the outstanding performance in a broad range of computer vision tasks~\cite{vit,carion2020end,chen2021pre,zheng2021rethinking,liu2021swin}, some methods have also investigated the potential of vision transformers in learned image compression (LIC). Bai~\emph{et al.}~\cite{bai2022towards} integrate LIC with ViT-based image analysis~\cite{vit} to exploit the synergy effect of these two tasks. Motivated by the window attention in Swin Transformer~\cite{liu2021swin}, Chen~\emph{et al.}~\cite{zou2022devil} present window attention-based symmetrical transformer (STF) framework which shows better R-D performance than previous CNN-based methods. In~\cite{Yichen_2022_ICLR}, a transformer-based entropy model is proposed to improve LIC by accuracy probability distribution estimation.

Beyond the above VAE style methods, there are also some other attempts for LIC. For example, Xie~\emph{et al.}~\cite{xie2021enhanced} propose an enhanced invertible encoding network based on invertible neural networks (INN)~\cite{laurent2015nice} to largely capture the lost information for better compression. Chang~\emph{et al.}~\cite{chang2022conceptual} present conceptual compression, which decomposes the image contents into structural and textural layers and separately compresses each of them, resulting in high reconstruction quality at very low bitrates. \cite{chang2022consistency} introduces consistency-contrast learning in conceptual compression to align the structure of texture representation space and source image space. \cite{GANcompression,dash2020compressnet} exploit the image generation capabilities of generative adversarial networks (GAN) to improve the perceptual quality of low-bitrate compressed images. Different from these methods, this work is inspired by the mask sampling and pre-training schemes in MAE and combines them with LIC to target effective extremely low-bitrate compression. 
\begin{figure*}[t]
  \centering
  \includegraphics[width=1.0\linewidth]{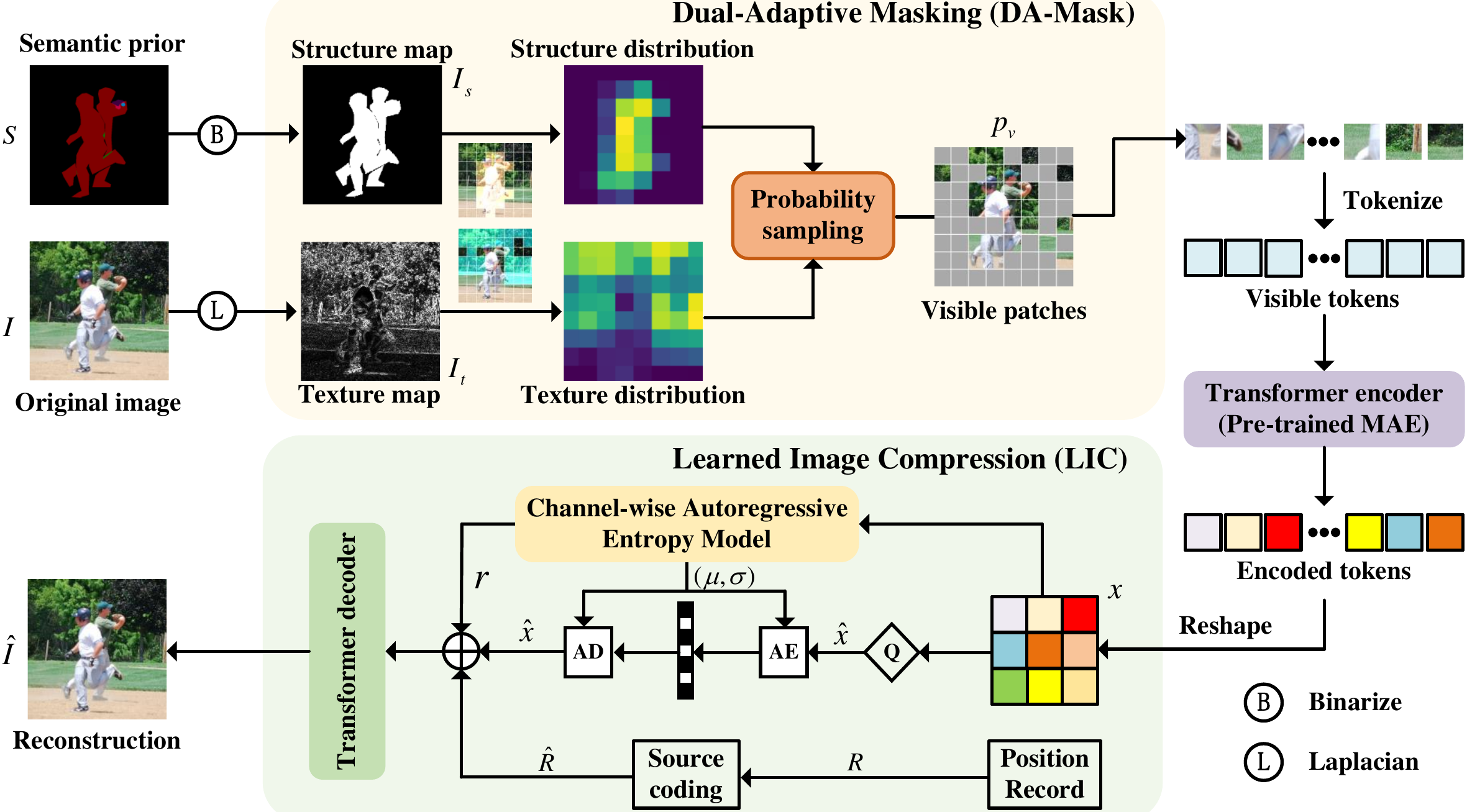}
  \caption{The overall framework of our masked compression model (MCM) which unifies pre-trained MAE~\cite{MAE}-based MIM and LIC for extremely low-bitrate image compression. We replace the random sampling in MAE with our dual-adaptive masking approach (DA-Mask). ``Q'', ``AE'', and ``AD'' denote quantization, arithmetic encoding, and arithmetic decoding, respectively.}
    \label{fig:framework}
\end{figure*}
\subsection{Masked Image Modeling}
Masked image modeling (MIM) can be derived from masked language modeling in natural language processing (NLP)~\cite{BERT,gpt1,gpt2,gpt3}, which adopts self-supervised pre-training paradigm to learn representations from unlabeled datasets by predicting the randomly masked patches in an image and then fine-tuning for specific downstream tasks. The representative method of MIM is masked autoencoder (MAE)~\cite{MAE} which applies a standard ViT~\cite{vit} as the encoder to operate on reserved visible patches and then reconstructs the missing patches with a lightweight decoder. 
In~\cite{SimMIM}, Xie~\emph{et al.} propose to regress the raw pixels of masked patches with $\ell_1$ loss, which performs comparably to patch classification approaches~\cite{vit,liu2021swin,BEiT}. CMAE~\cite{huang2022contrastive} imposes contrastive learning into MAE to improve the representation of MIM. GreenMAE~\cite{GreenMIM} extends the asymmetric encoder-decoder architecture of MAE to hierarchical vision transformers to better model hierarchical representations and local inductive bias. MCMAE~\cite{gao2022mcmae} introduces hybrid convolution-transformer architectures with masked convolution for mask autoencoding. 

In addition to the exploration of MIM on high-level vision tasks, there are also some works that have been proposed to study the effectiveness of MIM for image generation. Chang~\emph{et al.}~\cite{chang2022maskgit} propose a masked generative image transformer (MaskGIT) that learns to generate all tokens of an image from randomly masked tokens. MAGE~\cite{MAGE} unifies generative training and self-supervised representation learning in a single token-based MIM framework for class-unconditional image generation. Lezama~\emph{et al.}~\cite{lezama2022improved} propose a token-critic transformer to select which tokens should be accepted or resampled. Duan~\emph{et al.}~\cite{MAEIP} develop MAE pre-training based on an efficient window attention-based transformer to tackle image processing tasks such as image denoising, deblurring, and deraining \emph{et al.}.
However, the significance of MIM on image compression has not been researched. In this work, we show that the pre-trained MAE can also be seen as an initial compressor that can well cooperate with LIC architectures. Besides, we propose a dual-adaptive masking approach that samples visible patches based on the structure and texture distributions of original images, thus learning more informative representations than the above random sampling strategy in MIM.

\section{Proposed Method}\label{sec3}
\subsection{Problem Formulation}
The goal of our proposed masked compression model (MCM) is to provide a new paradigm for extremely low-bitrate compression. As illustrated in Fig.~\ref{fig:framework}, MCM is a two-stage framework that unifies pre-trained MAE~\cite{MAE}-based MIM and LIC in an end-to-end manner. Given an image $I$ and its corresponding semantic prior $S$, in the first stage, a dual-adaptive masking approach is utilized to perform adaptive patch sampling based on the structure and texture distributions of $I$. Here, the texture distribution of each patch is calculated over all patches according to the Laplacian-detected texture map $I_t$ and the structure distribution is obtained from the structure map $I_s$, \emph{i.e.} the binary version of the semantic map $S$. Thus, we can obtain texture- and structure-sensitive patches $p_t$ and $p_s$ which highly describe the appearance of visual objects. Then, probability sampling is utilized to decide the final candidate patches conditioned on $p_t$ and $p_s$. The main process of DA-Mask is formulated as:
\begin{equation}
    p_v = \mathcal{M}(I, S)
    \label{eq1}
\end{equation}
where $\mathcal{M}(\cdot)$ denotes the DA-Mask operation to obtain the visual patches $p_v$ in $I$.

After that, receiving the visible tokens by the tokenization on the visible patches $p_v$, similar to MAE~\cite{MAE}, we also apply a pre-trained standard ViT~\cite{vit} as the encoder $E$ to learn latent representations, 
resulting in encoded tokens $x$ of $p_v$:
\begin{equation}
    x = E(p_v)
    \label{eq2}
\end{equation}
Therefore, equipped with DA-Mask, our MCM can achieve initial compression from structure and texture perspectives, enabling to abstract more informative representations for later LIC.

In the second stage, we do not design specific architecture and only build the LIC module upon the channel-wise auto-regressive entropy model~\cite{minnen2020channel}. 
Especially, the encoded tokens $x$ are~first reshaped as the input of LIC for quantization $Q$, arithmetic encoding $AE$, arithmetic decoding $AD$, and entropy model.
Besides, different from existing LIC methods~\cite{minnen2020channel}, considering that the positional information of tokens is critical to locate the mask tokens attending in an image, we transmit the position record of visible patches $R$ with a source coding approach (Huffman coding~\cite{huffman1952method} in this work) to support for mask prediction. Finally, a lightweight transformer decoder $D$ is utilized to reconstruct the decompressed image $\hat{I}$. The main process of LIC stage can be formulated as:
\begin{equation}
\begin{aligned}
    \hat{x} &= Q(x) \\
    \hat{R} &= Huffman(R) \\
    \hat{I} &= D(\hat{x}, r, \hat{R})
    \label{eq3}
\end{aligned}
\end{equation}
where $Huffman$ denotes Huffman coding operations. $\hat{R}$ is the decoded position record of $R$. $r$ is the residual produced by the entropy model to reduce the quantization errors $x-\hat{x}$. 
\begin{figure}[t]
\flushright
\hfill
\begin{overpic}[width=\linewidth]{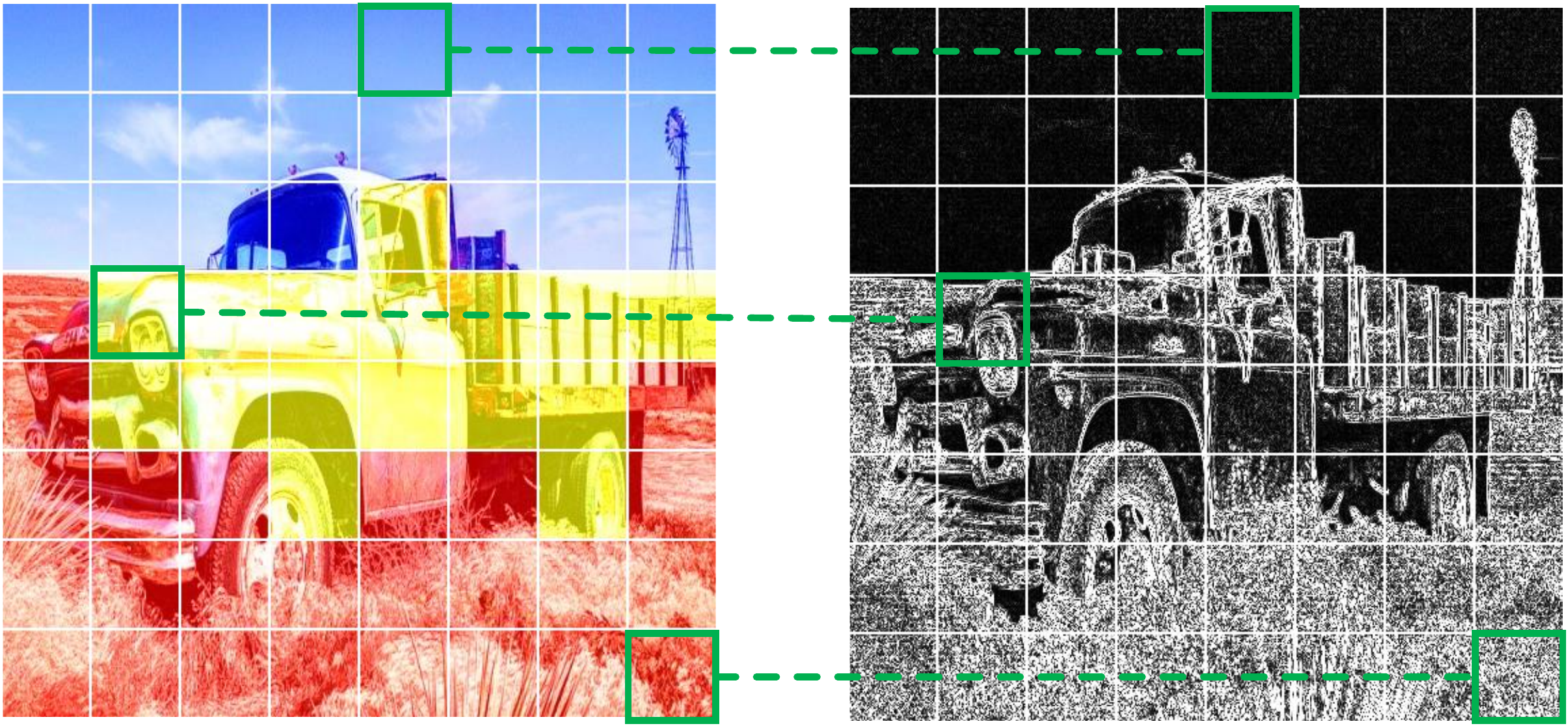}
\end{overpic}\hfill
\begin{overpic}[tics=25,width=\linewidth]{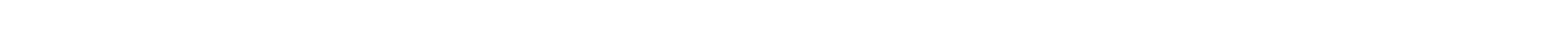}
\put(21,0){\textbf{Original image}}
\put(155,0){\textbf{Texture map}}
\end{overpic}\hfill
\begin{overpic}[tics=25,width=\linewidth]{figures/Qualitative/0.pdf}
\end{overpic}\hfill
\begin{overpic}[tics=25,width=0.93\linewidth]{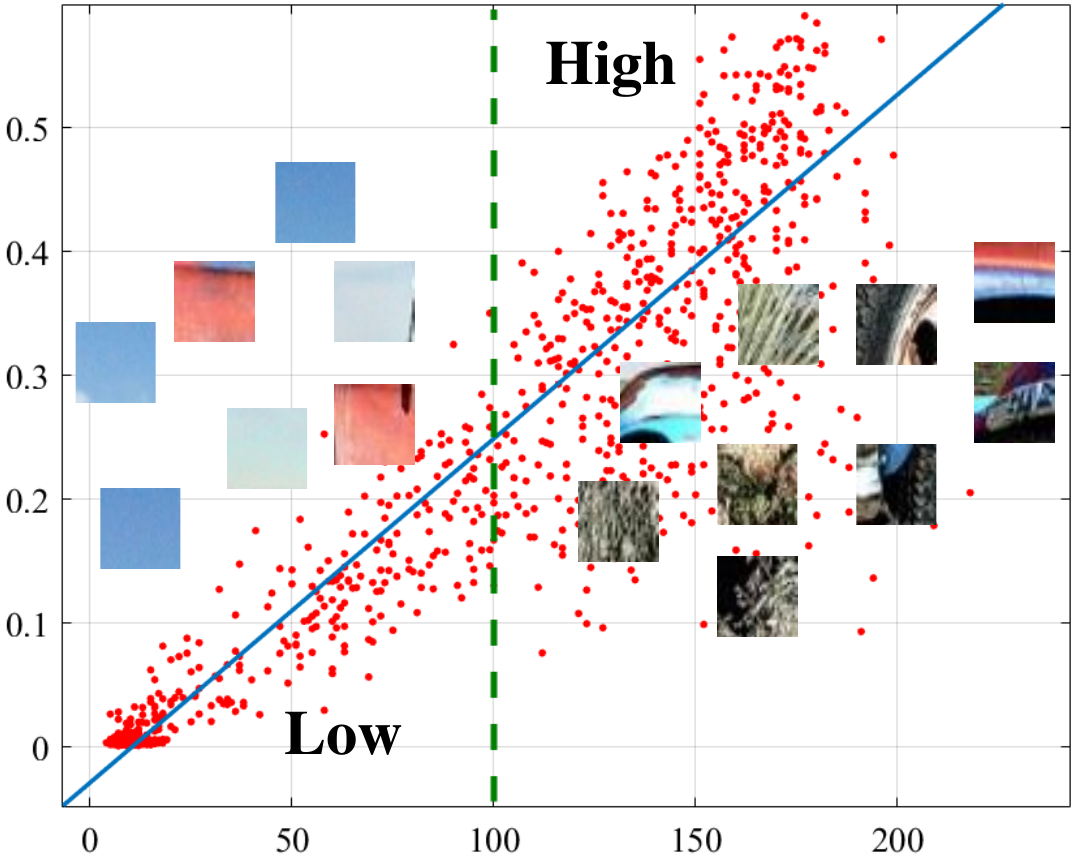}
\end{overpic}\hfill
\begin{overpic}[tics=25,width=\linewidth]{figures/Qualitative/0.pdf}
\put(80,0){\textbf{Texture Complexity Score}}
\put(7,90){\begin{turn}{90}\textbf{LPIPS}\end{turn}}
\end{overpic}
\caption{The correlation between texture complexity and reconstruction quality. Top: the original image and detected texture map by 2nd-order Laplacian differential operator. Bottom: the LPIPS of the image patches with different texture complexity scores.}
\label{fig:edge_lpips}
\end{figure}
\subsection{Dual-Adaptive Masking}
Since different patches contain different information, simply implementing
such a random mask sampling as existing MIM methods~\cite{MAE,chang2022maskgit,MAGE} may preserve some meaningless patches and discard some important texture or structure information that is important for visual understanding, resulting in misleading reconstruction. Therefore, we propose the dual-adaptive masking (DA-Mask) approach to adaptively sample informative patches that are conducive to image reconstruction while achieving effective redundancy removal. As shown in Fig.~\ref{fig:framework}, DA-Mask contains two branches that model the texture and structure complexity of all patches to guide later probability sampling.

\textbf{Texture Complexity Modeling}. Considering that the edge prior reveals the sharpness of local regions in an image and is high-related to texture details recovery, we quantify the edge information of each patch to measure the texture complexity. To be specific, given an RGB image $I$ with the spatial size of $H\times W$, the texture map $I_t$ is obtained by the edge information extraction on $I$ using a 2nd-order Laplacian differential operator which has been demonstrated to have a more stable location ability, better sharpness, and more robust to noise~\cite{sun2008image}
\begin{gather}
I_{g} = RGB2Gray(I) \\
I_{t} = \frac{\partial^{2}I_{g}}{\partial x^{2}} + \frac{\partial^{2}I_{g}}{\partial y^{2}},\;I_{t} \in \mathbb{R}^{1\times H \times W}
\end{gather}
where $I_{g}$ is the grayscale version of $I$ produced by the convert function $RGB2Gray$. We then split both $I$ and $I_t$ into serial non-overlapped patches, denoted as $p \in \mathbb{R}^{L}$ and $p_t\in \mathbb{R}^{L}$, where $L = \frac{H}{N} \, \times \, \frac{W}{N}$ represents the number of patches with the spatial size of $N \times N$. The texture complexity score of a certain image patch is calculated by absolutely summing the absolute values of all elements inside its corresponding texture patch. 
\begin{equation}
Score^l_t = \sum_{i=1}^{N} \sum_{j=1}^{N} |p_t^l(i,j)|
\label{eq6}
\end{equation}
where $Score^l_t$ is the score of the $l$-th patch $p^l$. $(i,j)$ denotes the position of the patch. In this work, we set the patch size $N\times N$ as $16\times 16$. Therefore, for $L$ patches, we can get all the scores $(Score^1_t,Score^2_t,...,Score^L_t)$. 

The correlation between texture complexity and reconstruction quality is illustrated in Fig.~\ref{fig:edge_lpips}. We adopt BPG~\cite{BPG} to compress them and rank their performance using Learned Perceptual Image Patch Similarity (LPIPS). It can be observed that the patches with low texture complexity tend to have low LPIPS (better), while the patches with high LPIPS (worse) contain more textures. Based on this observation, we can conclude that: 1) The patch involving large texture complexity is hard to be reconstructed after compression; 2) Masking a portion of very smooth patches yet preserving complex ones can effectively reduce the pixel redundancy and alleviate the prediction burden on local textures. 
\begin{figure}[t]
\hfill
\begin{overpic}[width=\linewidth]{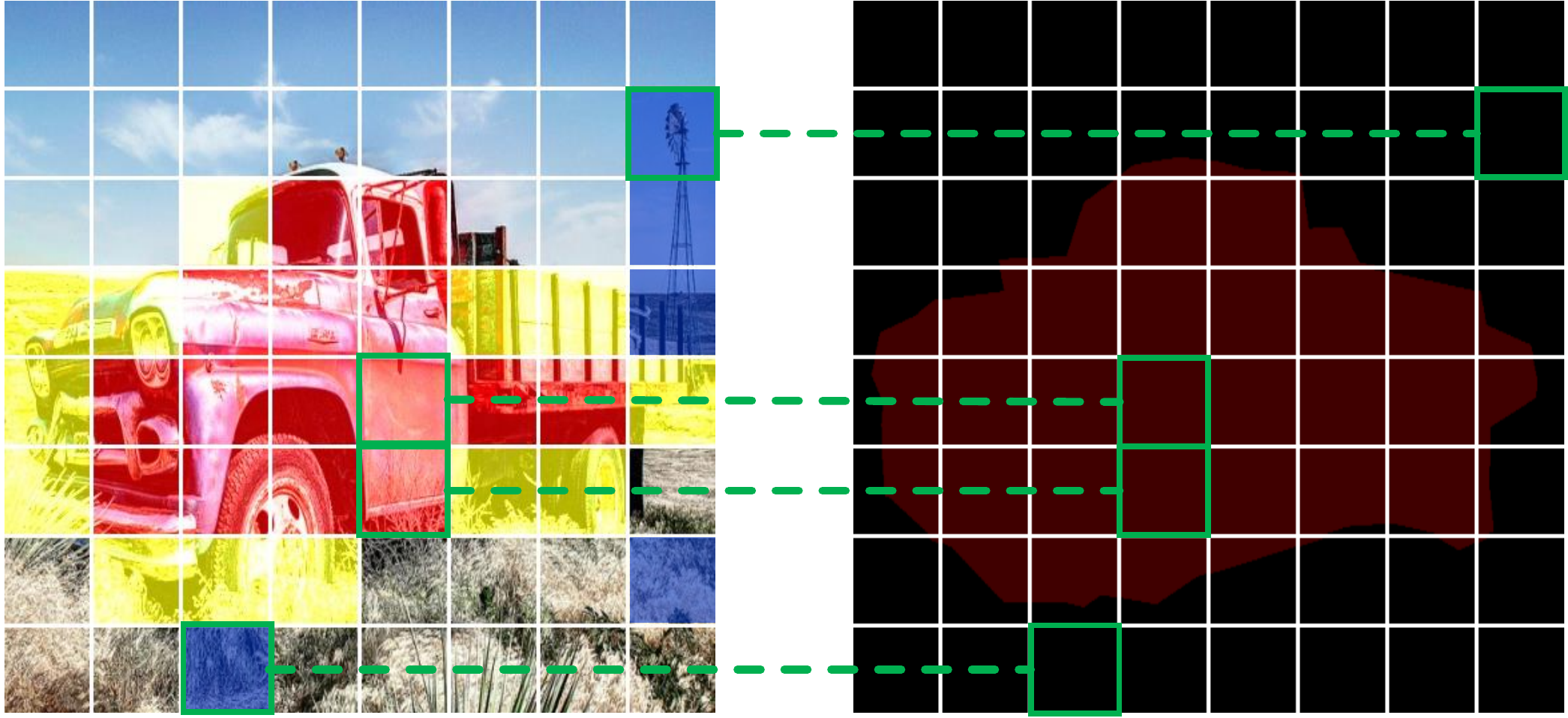}
\end{overpic}\hfill
\begin{overpic}[tics=25,width=\linewidth]{figures/Qualitative/0.pdf}
\put(21,0){\textbf{Original image}}
\put(155,0){\textbf{Semantic map}}
\end{overpic}\hfill
\caption{The patch-wise correspondences between the original image and semantic map.}
\label{fig:semantic}
\end{figure}

\textbf{Structure Complexity Modeling}. Though the texture-based patch sampling strategy can help to learn texture-aware representations, as shown in Fig. \ref{fig:semantic}, it can lead to the network over-emphasizing the background regions with complex textures ({\color{blue}{blue patches}}) and ignoring some important smooth parts ({\color{red}{red patches}}) of visual objects. In fact, it is crucial to capture the semantic context of objects for both visual analysis and image synthesis tasks. To address this problem, we propose to model the structure complexity as an additional guidance for more informative patch sampling. 

Specifically, for the input image $I$, we adopt its semantic map $S$ as prior to generate a structure map $I_s$ by binary operation. Similar to the texture-based strategy, besides the image patch $p$, we also split $I_s$ into multiple $N\times N$ non-overlapped patches, resulting in $p_s$. Then, we measure the structure complexity by directly summing all the internal elements 
\begin{equation}
Score^l_s = \sum_{i=1}^{N} \sum_{j=1}^{N} p_s^l(i,j)
\label{eq8}
\end{equation}
where $Score^l_s$ denotes the semantic score of the $l$-th patch. Correspondingly, for total $L$ patches of $I$, we can obtain the scores $(Score^l_1,Score^l_2,...,Score^L_s)$. In this way, we can effectively capture the semantic richness of each patch that describes the global structure of visual objects, which helps DA-Mask perform semantic-oriented masking.

\begin{figure}[t]
\centering
\hfill
\begin{overpic}[width=\linewidth]{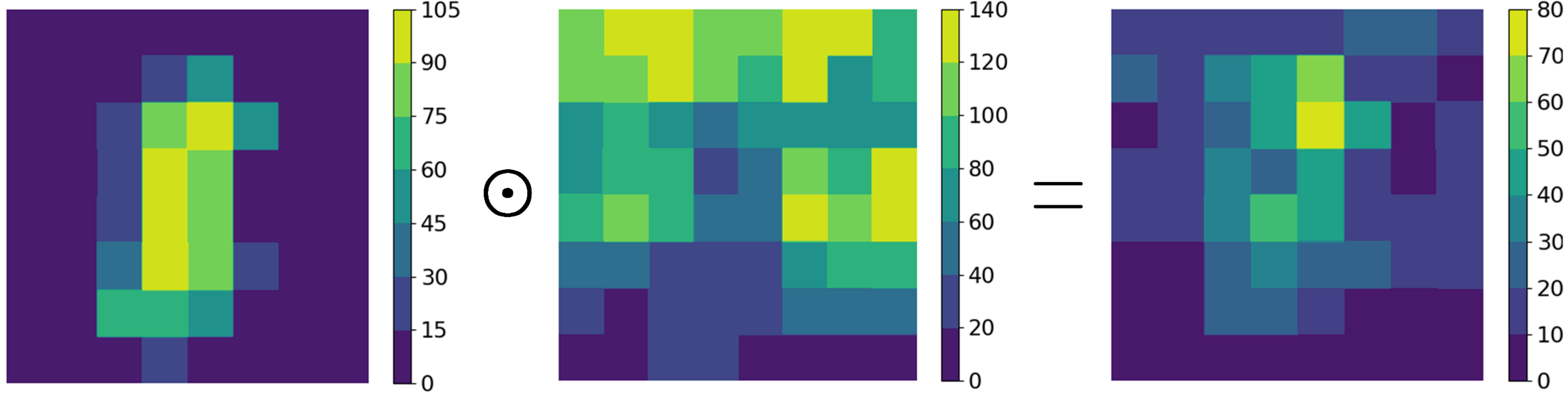}
\end{overpic}\hfill
\begin{overpic}[tics=25,width=\linewidth]{figures/Qualitative/0.pdf}
\put(10,0){Structure}
\put(97,0){Texture}
\put(176,0){Information}
\end{overpic}\hfill
\begin{overpic}[tics=25,width=\linewidth]{figures/Qualitative/0.pdf}
\put(5,0){distribution}
\put(90,0){distribution}
\put(176,0){distribution}
\end{overpic}
\caption{The visualization of structure, texture, and information distributions based on corresponding complexity scores and information capacity.}
\label{fig:distribution}
\end{figure}
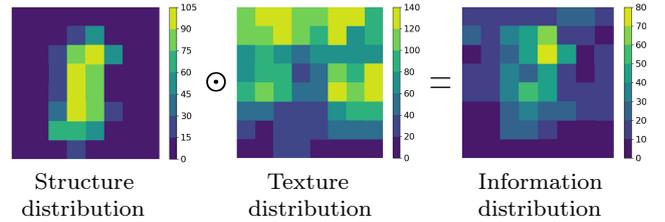

\textbf{Probability Sampling}. As mentioned above, both the texture and structure complexity of each patch are important for informative representation learning and content recovery.
Therefore, in DA-Mask, probability sampling is introduced to select visible patches based on the texture and structure distributions of input image $I$. As shown in Fig.~\ref{fig:distribution}, we generalize the texture and structure complexity scores of all the patches to represent the overall texture and structure distributions respectively. After that, we conduct patch-wise multiply on the texture and structure complexity scores as the information capacity of each patch to reveal which one is beneficial for discriminative learning or meaningless.
\begin{equation}
    Inf_l = Score^l_s \odot Score^l_t
    \label{eq9}
\end{equation}
where $Inf_l$ denotes the information capacity of the $l$-th patch. A toy example of resulted information distribution is shown in Fig.~\ref{fig:distribution}. We then estimate the categorical distribution $Cat$ over all image patches according to the softmax normalized $Inf_l$.
\begin{equation}
Cat(\alpha_1,\alpha_2,..., \alpha_L) = Softmax(Inf_l), \;\sum_{l}^{L} \alpha_l = 1\\
\label{eq}
\end{equation}
where $(\alpha_1,\alpha_2,..., \alpha_L)$ denote the normalized parameters of $(Inf_1,Inf_2,...,Inf_L)$. Supposing the masking ratio is $\rho\in(0,1)$, the number of masked patches is equal to $L\times \rho$. Gumbel-softmax trick~\cite{jang2017categorical} is leveraged to determine the visible patch set $p_v$ which contains $L\times(1-\rho)$ sampled patches. 
In summary, based on the above analysis, our DA-Mask enables sampling the patches that can well balance the local texture and structural semantics and is prone to mask the patches with less information, resulting in effective redundancy removal. 
\subsection{Transformer Encoder/Decoder}\label{subsec3}
\begin{figure}
  \centering
  \includegraphics[width=1.0\linewidth]{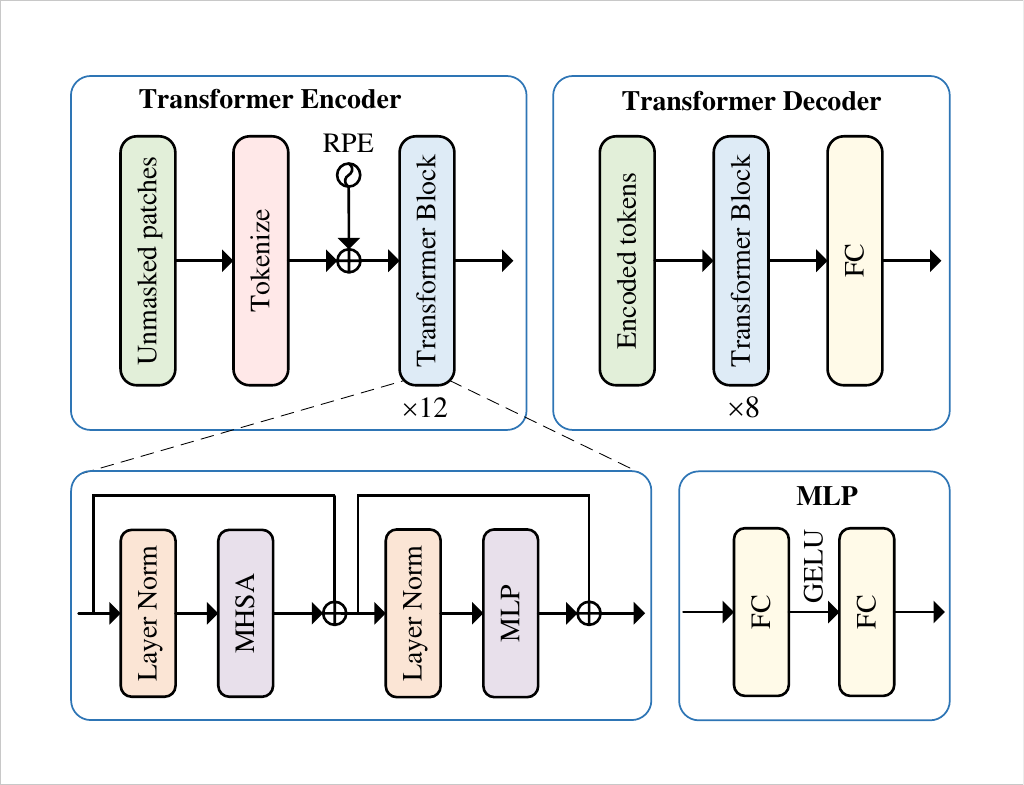}
  \caption{Architectures of the transformer encoder and transformer decoder in our MCM. Here, ``RPE'' and ``FC'' denote relative position embedding and full-connected layer.}
    \label{fig:encoder_decoder}
\end{figure}

Following~\cite{MAE}, as shown in Fig. \ref{fig:encoder_decoder}, for the encoder, we use the standard ViT with relative position embedding as the backbone which contains cascaded 12 transformer blocks. Each block consists of a multi-head self-attention (MHSA) and a multi-layer perceptron (MLP) with layer normalization (Layer Norm). 
The transformer decoder is more lightweight, which uses 8 transformer blocks and an additional full-connected layer followed by reshape operation to reconstruct the final image. 
\subsection{Loss Function}
As shown in Fig.\ref{fig:framework}, our MCM unifies pre-trained MAE~\cite{MAE}-based MIM and LIC for extremely low-bitrate image compression. Therefore, it involves two training stages, \emph{i.e.}, the pre-training on our transformer encoder in MIM to learn meaningful representation, and the joint optimization of overall MCM. The former is trained on ImageNet-1K dataset using MSE as the loss function to constrain the original image $I$ and reconstructed image $\hat{I}$. In the second stage, the LIC module receives initialized features by the pre-trained weights and then is jointly optimized with the MIM using Rate-Distortion (R-D) loss $\mathcal{L}$:
\begin{equation}
\mathcal{L} = \mathcal{R} + \lambda \mathcal{D}(I, \hat{I}),
\label{equ:RD_loss}
\end{equation}
where $\lambda$ is a Lagrangian multiplier factor to control the R-D trade-off. $\mathcal{R}$ denotes the bitrates estimated by the entropy model. 
The distortion loss $\mathcal{D}$ is a composited function that contains $L_1$, SSIM~\cite{wang2004image}, and VGG~\cite{johnson2016perceptual} losses, denoted by 
\begin{equation}
\mathcal{D} = \lambda_1 \Vert I - \hat{I} \Vert_1 
+ \lambda_2 SSIM(I, \hat{I}) + 
\lambda_3 VGG(I, \hat{I}),
\label{equ:distortion_loss}
\end{equation}
where $\lambda_1$, $\lambda_2$ and $\lambda_3$ are hyper-parameters that balance the three components. In this work, we set them to 10, 0.25, and 0.1 respectively.
\section{Experiments}
\begin{figure*}[t]
\centering
\begin{minipage}[t]{0.49\linewidth}
\flushright
\includegraphics[width=\linewidth]{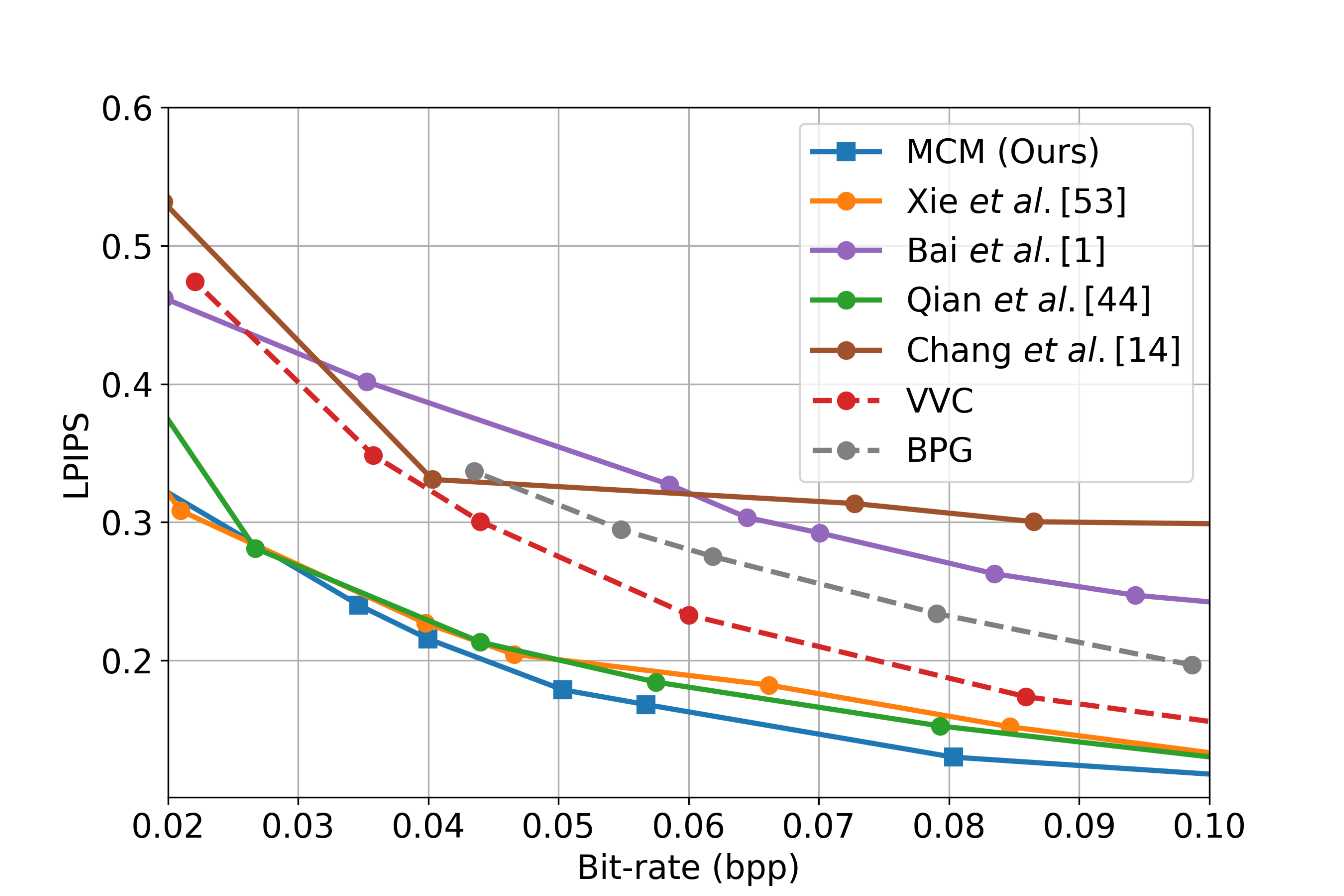}
\end{minipage}
\hfill
\begin{minipage}[t]{0.49\linewidth}
\flushleft
\includegraphics[width=\linewidth]{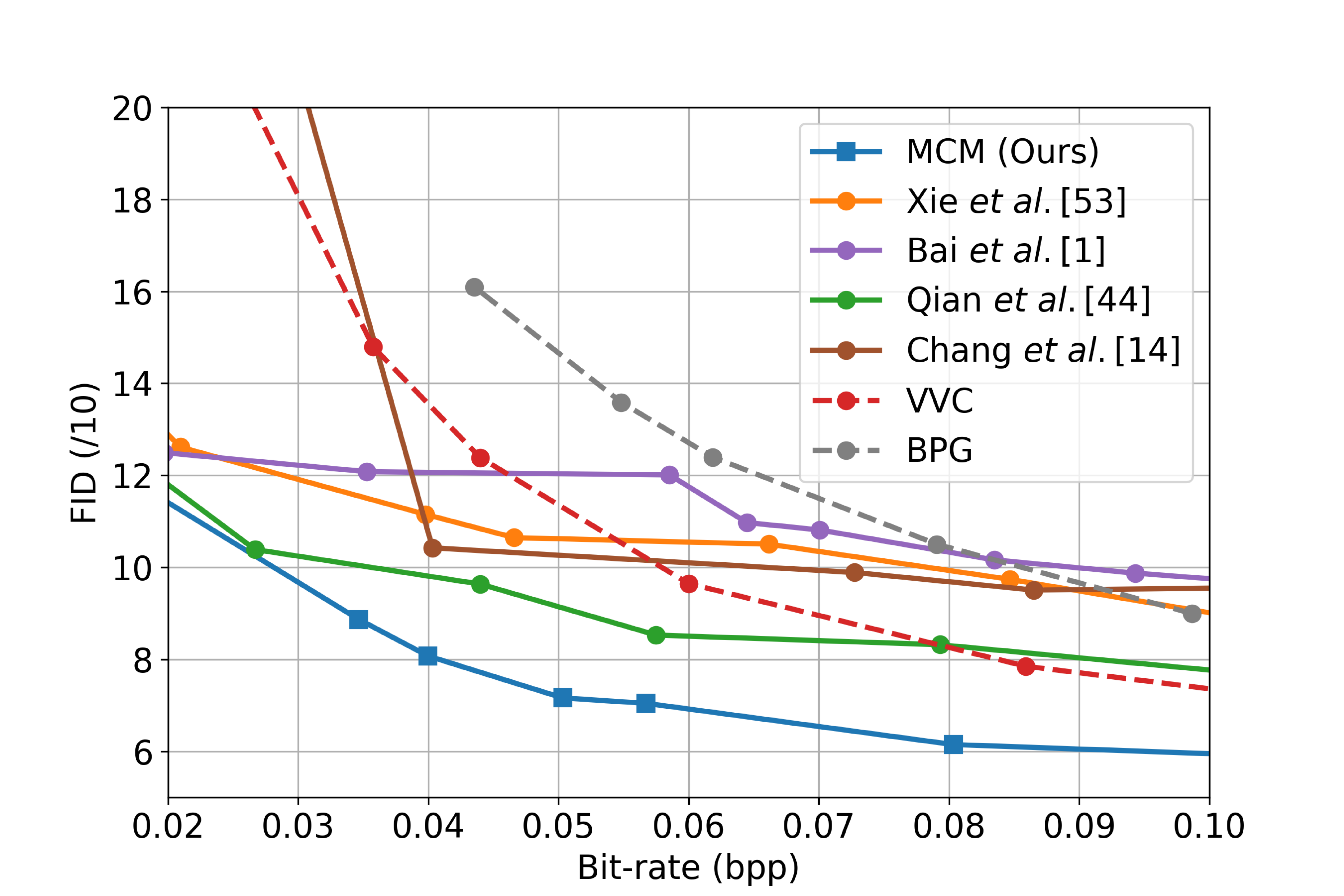}
\end{minipage}
\caption{The Rate-LPIPS and Rate-FID comparison on CelebAMask-HQ dataset.}
\label{fig:CelebAMask-HQ}
\end{figure*}
\begin{figure*}
\centering
\begin{minipage}[t]{0.49\linewidth}
\flushright
\includegraphics[width=\linewidth]{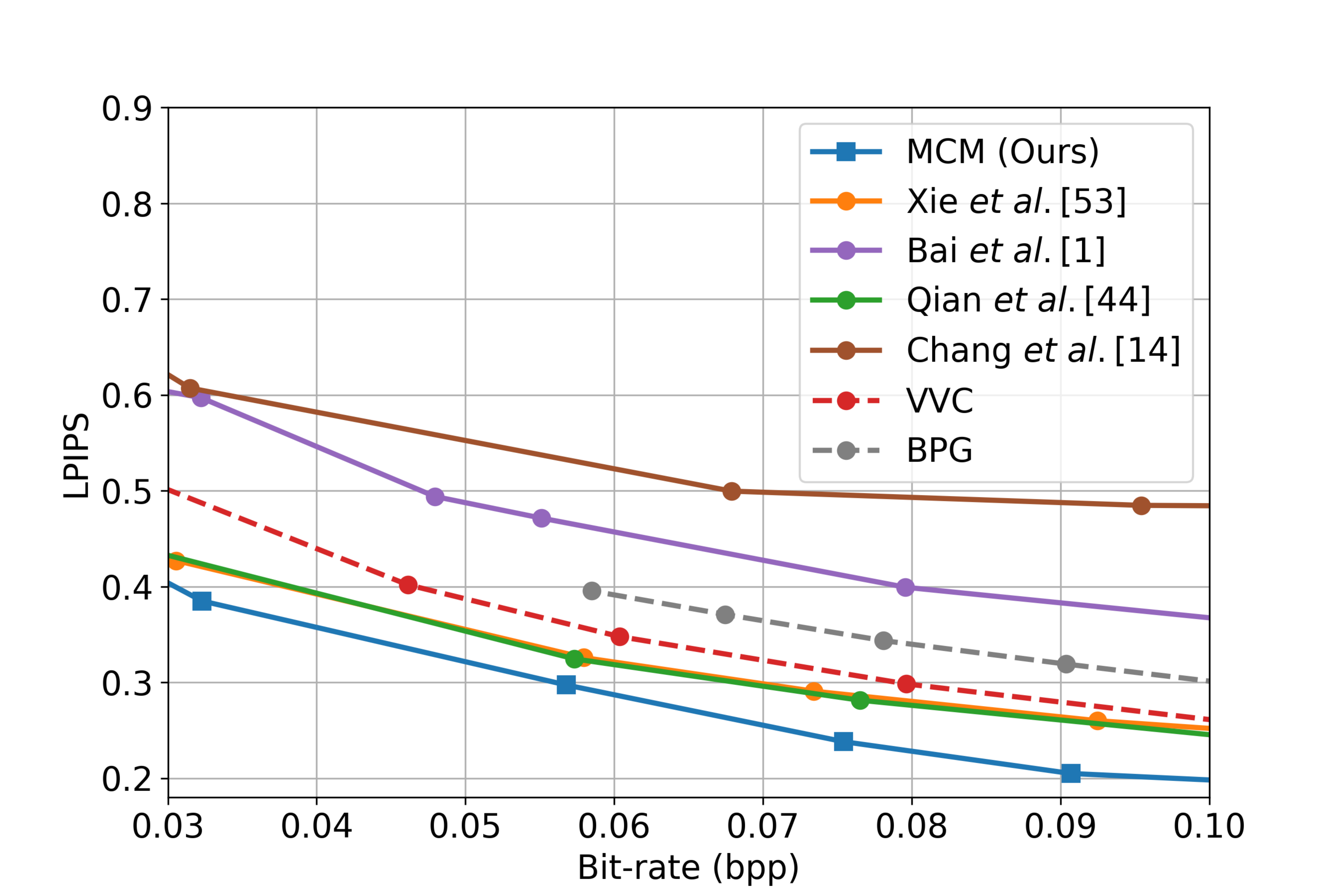}
\end{minipage}
\hfill
\begin{minipage}[t]{0.49\linewidth}
\flushleft
\includegraphics[width=\linewidth]{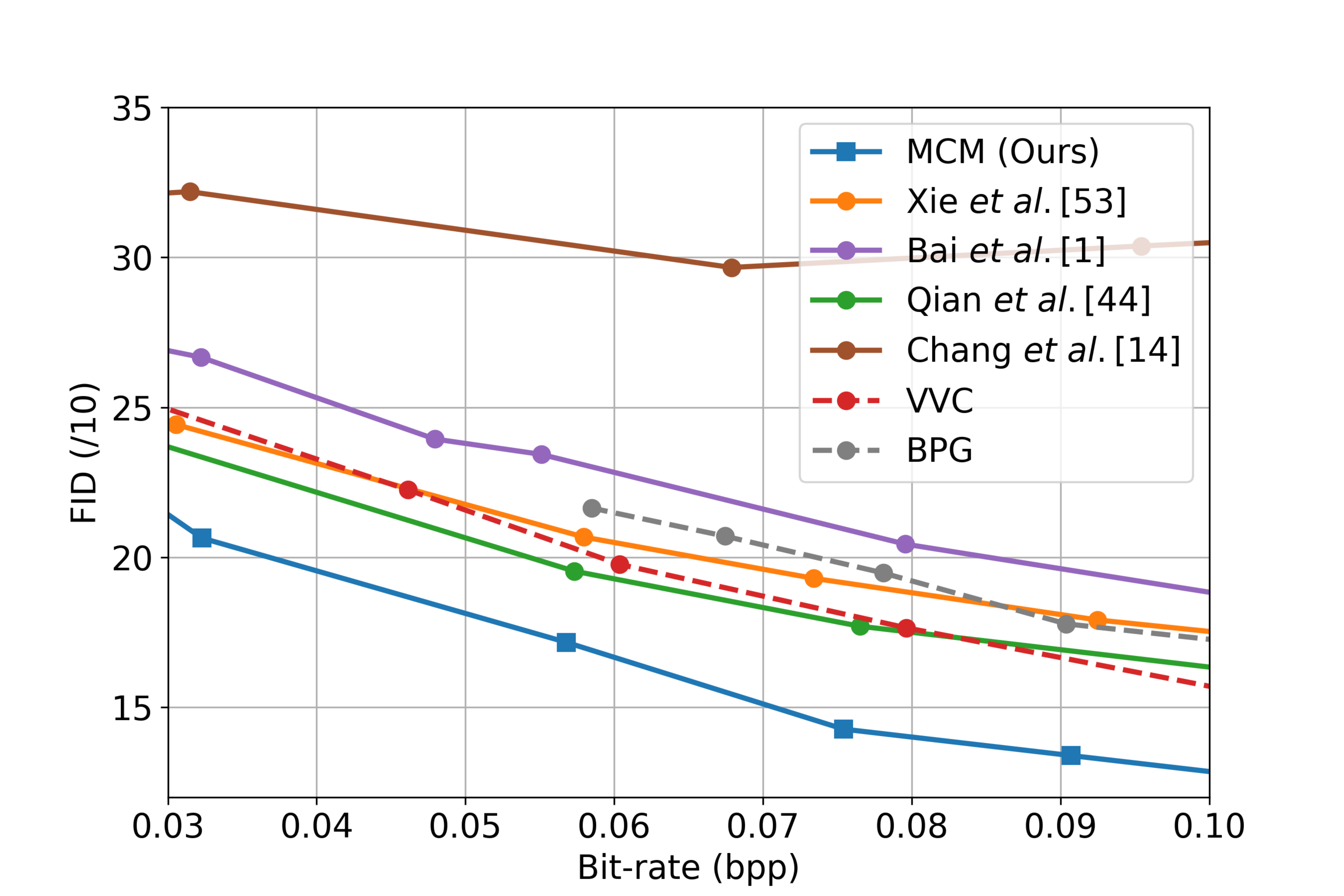}
\end{minipage}
\caption{The Rate-LPIPS and Rate-FID comparison on COCO dataset.}
\label{fig:COCO}
\end{figure*}
\begin{figure*}
\hfill
\begin{overpic}[tics=20,width=\textwidth]{figures/Qualitative/0.pdf}
\put(22,4){Original}
\put(110,4){VVC}
\put(173,4){Chang ${et}$ ${al.}$ \cite{chang2022conceptual}}
\put(260,4){Xie ${et}$ ${al.}$ \cite{xie2021enhanced}}
\put(341,4){Qian ${et}$ ${al.}$ \cite{Yichen_2022_ICLR}}
\put(429,4){MCM (Ours)}
\end{overpic}\hfill
\begin{overpic}[width=\textwidth]{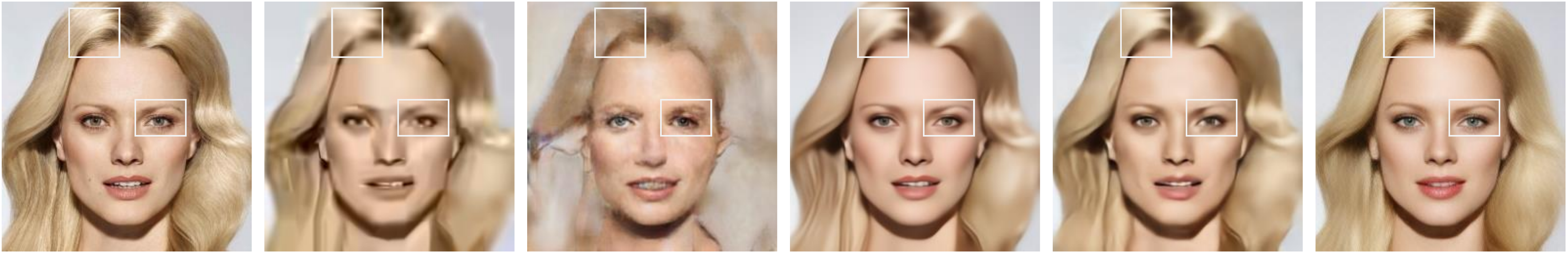}
\put(20,20){}
\end{overpic}\hfill
\begin{overpic}[tics=25,width=\textwidth]{figures/Qualitative/0.pdf}
\put(96,5){0.0617/0.252}
\put(180,5){0.062/0.280}
\put(268,5){0.062/0.166}
\put(350,5){0.066/0.175}
\put(429,5){\textbf{0.059/0.113}}
\end{overpic}\hfill
\begin{overpic}[width=\textwidth]{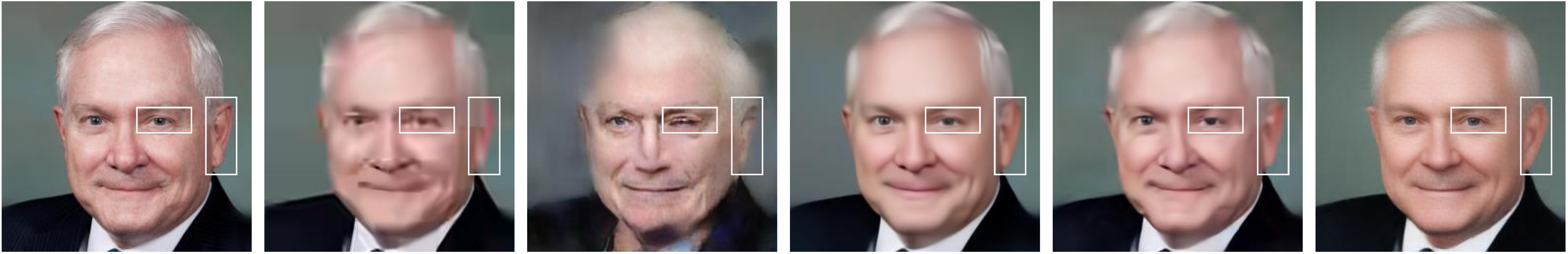}
\put(20,20){}
\end{overpic}\hfill
\begin{overpic}[tics=25,width=\textwidth]{figures/Qualitative/0.pdf}
\put(96,5){0.048/0.242}
\put(180,5){0.051/0.250}
\put(268,5){0.045/0.145}
\put(350,5){0.054/0.162}
\put(429,5){\textbf{0.047/0.110}}
\end{overpic}
\begin{overpic}[width=\textwidth]{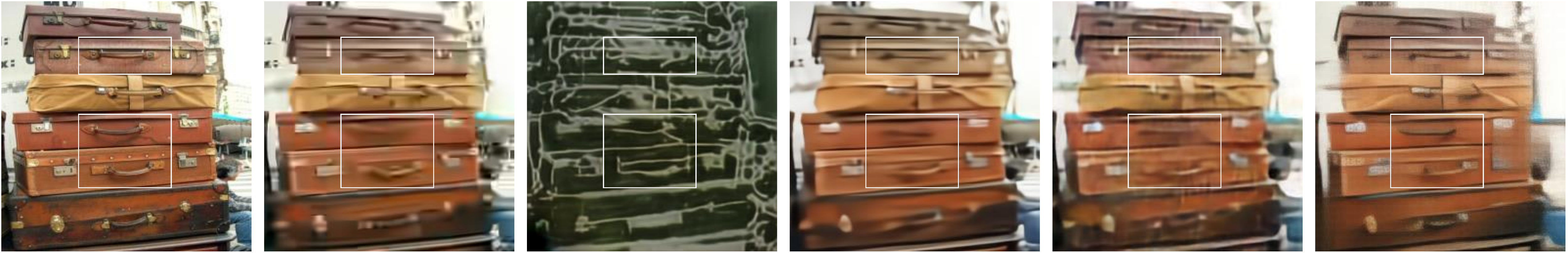}
\put(10,10){}
\end{overpic}\hfill
\begin{overpic}[tics=25,width=\textwidth]{figures/Qualitative/0.pdf}
\put(96,5){0.106/0.278}
\put(180,5){0.089/0.556}
\put(268,5){0.092/0.298}
\put(350,5){0.089/0.256}
\put(429,5){\textbf{0.085/0.245}}
\end{overpic}\hfill
\begin{overpic}[width=\textwidth]{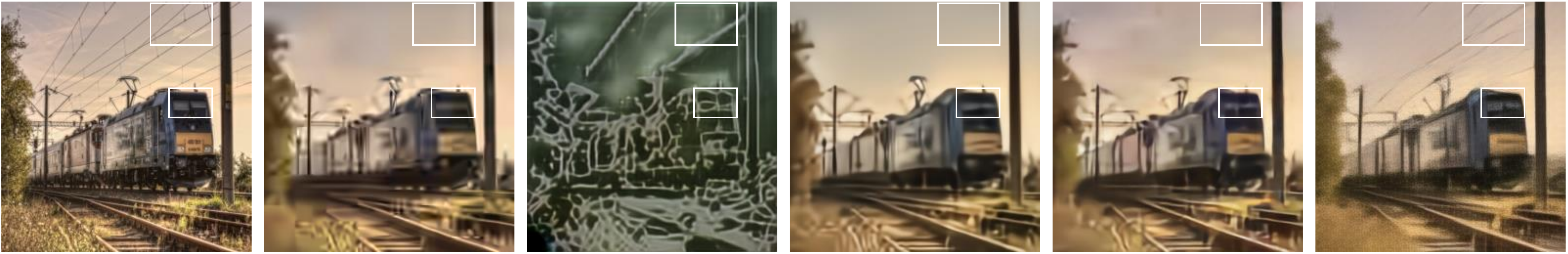}
\put(20,20){}
\end{overpic}\hfill
\begin{overpic}[tics=25,width=\textwidth]{figures/Qualitative/0.pdf}
\put(96,5){0.109/0.148}
\put(180,5){0.094/0.535}
\put(268,5){0.093/0.135}
\put(350,5){0.102/0.122}
\put(429,5){\textbf{0.091/0.134}}
\end{overpic}\hfill
\begin{overpic}[width=\textwidth]{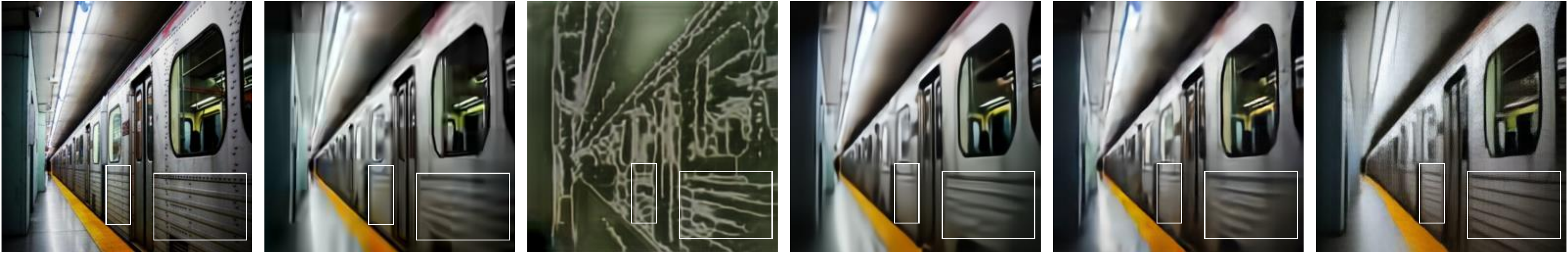}
\put(20,20){}
\end{overpic}\hfill
\begin{overpic}[tics=25,width=\textwidth]{figures/Qualitative/0.pdf}
\put(96,5){0.091/0.211}
\put(180,5){0.093/0.507}
\put(268,5){0.091/0.207}
\put(350,5){0.094/0.209}
\put(429,5){\textbf{0.086/0.209}}
\end{overpic}
\caption{Visual performance on CelebAMask-HQ and COCO dataset. Bpp$\downarrow$/LPIPS$\downarrow$ of each method are also shown for comparison.}
\vspace{-0.25cm}
\label{fig:Qualitative}
\end{figure*}
\begin{figure}[t]
\hfill
\begin{overpic}[tics=20,width=0.585\linewidth]{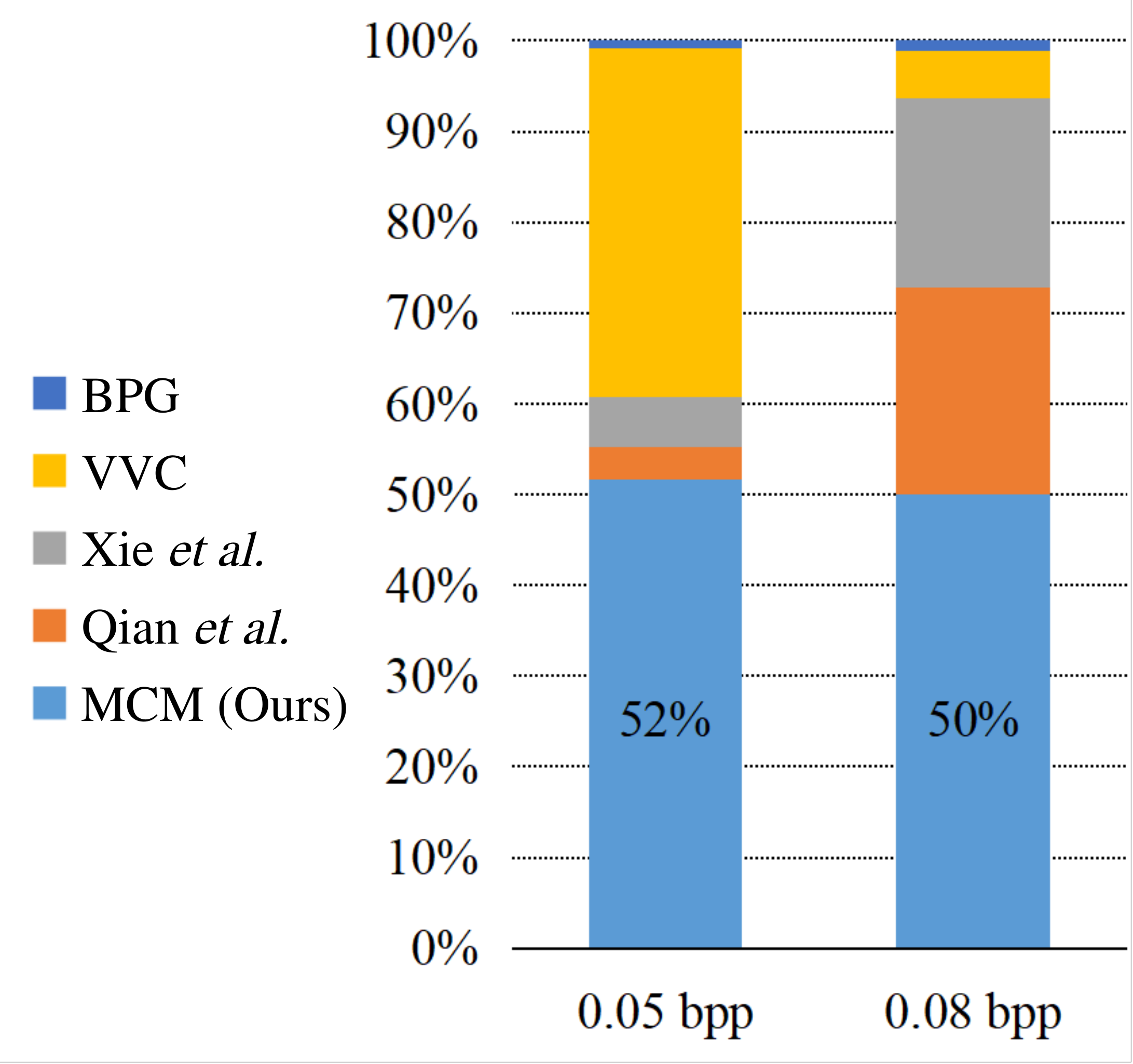}
\end{overpic}
\begin{overpic}[width=0.41\linewidth]{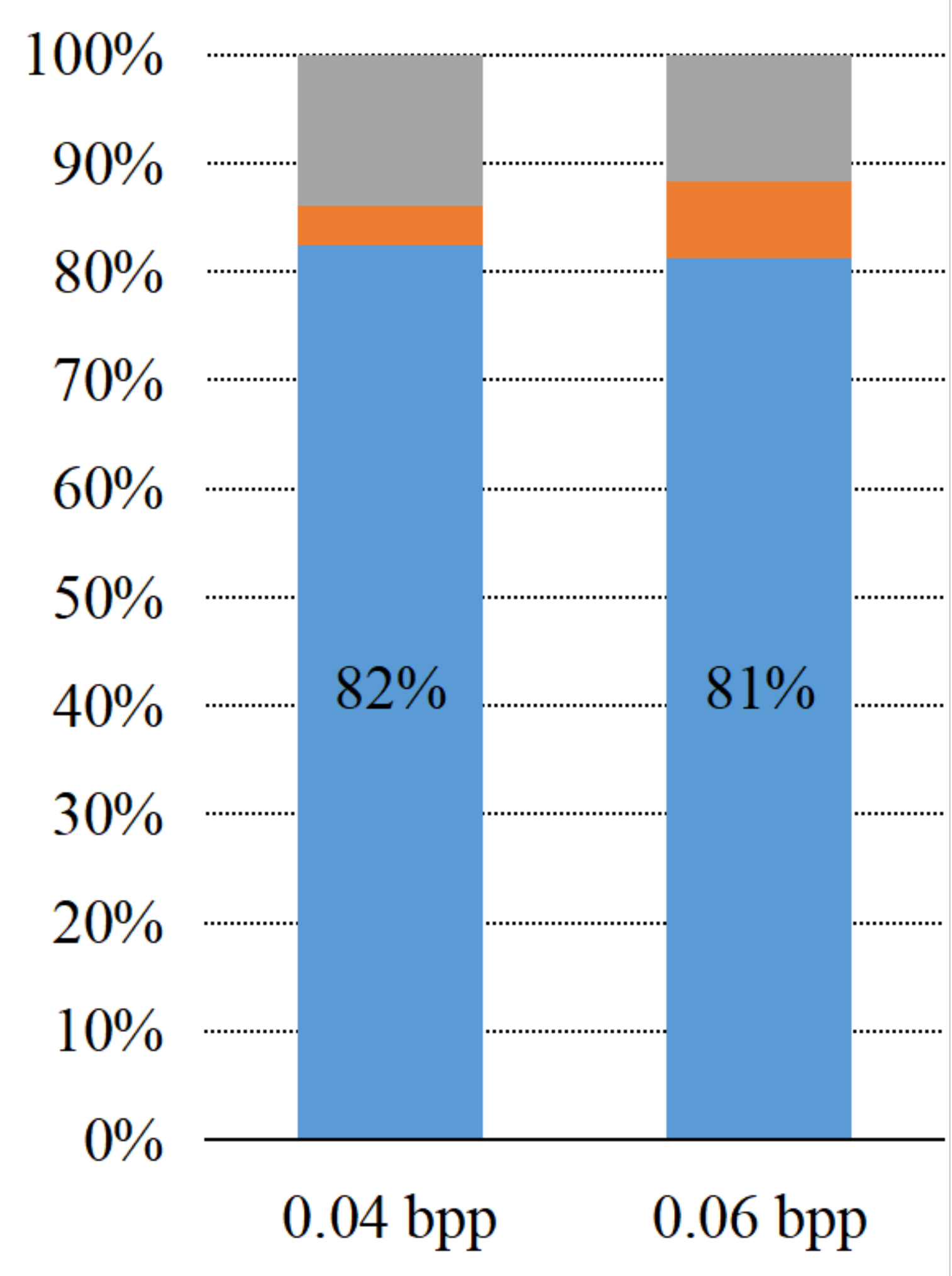}
\end{overpic}\hfill
\begin{overpic}[tics=25,width=\linewidth]{figures/Qualitative/0.pdf}
\put(96,0){(a)}
\put(197,0){(b)}
\end{overpic}
\caption{The human opinion of various methods and ours on COCO (Fig(a)) and CelebAMask-HQ (Fig(b)) validation dataset.}
\label{fig:user_study}
\end{figure}
\subsection{Experimental Setup}
We use MS COCO~\cite{lin2014microsoft} and CelebAMask-HQ~\cite{lee2020maskgan} to evaluate the performance of our MCM. Here, MS COCO contains 82783 images for training and 40504 images for testing. We randomly select 200 images from the testing set for performance evaluation. CelebAMask-HQ contains 30000 high-resolution face images and we randomly split 29800 images for training and 200 images for testing. All images in both datasets are resized to $256\,\times\,256$ as the input. 
For pre-training, we train the Transformer encoder and Transformer decoder using ImageNet-1K where the mask ratio is set to 75\% for all images. Other training details, such as learning rate, optimizer, and weight decay are consistent with the setting in \cite{MAE}. 
Then, the overall MCM is trained for 250K iterations on COCO and 90K iterations on CelebAMask-HQ. 
All the experiments are implemented on NVIDIA RTX 3090 GPUs. 
\subsection{Comparison with The State-of-the-Arts}
We compare our MCM with recent state-of-the-art
(SOTA) learned end-to-end image compression methods, including Xie~\emph{et al.}~\cite{xie2021enhanced}, Qian~\emph{et al.}~\cite{Yichen_2022_ICLR}, Bai~\emph{et al.}~\cite{bai2022towards} and Chang~\emph{et al.}~\cite{chang2022conceptual} as well as traditional image compression methods BPG~\cite{BPG} and VVC~\cite{bross2021developments} on extremely low-bitrate conditions ($<0.1$bpp). For a fair comparison, all the LIC methods are re-trained using their public source codes on the same datasets as ours. 
For BPG, we use the BPG software with YUV444 subsampling, x265 HEVC implementation, and 8-bit depth. For VVC, we use the VVC Official Test Model VTM 10.0 with an intra-profile configuration. 

\textbf{Rate-distortion Performance.}~
Because the reconstructed images are heavily distorted at very low bitrates, pixel-level measurements such as classical PSNR and MS-SSIM can not well reflect the reconstruction quality. Therefore, in this work, we use Learned Perceptual Image Patch Similarity (LPIPS)~\cite{zhang2018unreasonable} and Fr{\'e}chet inception distance (FID)~\cite{heusel2017gans} which are widely used in image generation tasks~\cite{rombach2022high,chang2019free,EdgeConnect}, as the metrics to separately evaluate the perceptual quality of images and the distance of the original data and reconstructed data. Fig. \ref{fig:CelebAMask-HQ} and Fig. \ref{fig:COCO} illustrate the Rate-LPIPS and Rate-FID on CelebAMask-HQ and COCO, respectively. Firstly, in Fig. \ref{fig:CelebAMask-HQ}, compared to existing LIC methods, classical coding standards BPG~\cite{BPG} and VVC~\cite{bross2021developments} tend to perform better at higher bitrates (0.08-0.1bpp). Chang~\emph{et al.}~\cite{chang2022conceptual} adopts a conceptual coding to compress structure and texture components individually for low bitrate, which shows worse R-D performance than other compared methods in terms of Rate-LPIPS but better than Xie~\emph{et al.}~\cite{xie2021enhanced} and Bai~\emph{et al.}~\cite{bai2022towards} on Rate-FID. Though Qian~\emph{et al.}~\cite{Yichen_2022_ICLR} outperforms existing LIC methods and VVC~\cite{bross2021developments} on both metrics, it is still inferior against our method at almost every R-D point.
In Fig. \ref{fig:COCO}, for the more challenging dataset COCO, we can see that the performance of Chang~\emph{et al.}~\cite{chang2022conceptual} deteriorates heavily, no matter LPIPS or FID. By comparison, our method achieves the SOTA performance, which surpasses other methods by a large margin, especially on Rate-FID. 

\textbf{Qualitative Comparison.}~
In Fig. \ref{fig:Qualitative}, we illustrate the visual results of all methods for qualitative comparison.
Generally, it can be observed our method can reconstruct the images with a more vivid texture appearance and global structural fidelity at low bitrates, while other methods produce results that suffer from more blurs and fewer local details with higher bitrates. As for CelebAMask-HQ, although all the compared methods show good visual quality, our results contain more details in the key facial components, such as the eyes and mouth. 

\textbf{BD-rate savings.}~
\begin{table}[t]
\caption{Average BD-rate savings on COCO Dataset and CelebAMask-HQ Dataset. We adopt VVC as the anchor.}
\centering
\begin{tabular}{lcc}
\toprule[1pt]
\multirow{1}{*}{Method} & COCO & CelebAMask-HQ \\ \hline
BPG  & 26.56\% & 26.39\% \\
Bai~\emph{et al.}~\cite{bai2022towards} & 65.94\% & 33.50\%  \\
Chang~\emph{et al.}~\cite{chang2022conceptual} & 162.96\% & 1.23\%  \\
Qian~\emph{et al.}~\cite{Yichen_2022_ICLR} & -20.16\% & -38.68\%  \\
Xie~\emph{et al.}~\cite{xie2021enhanced}  & -18.51\% & -35.96\% \\
MCM (Ours) & \textbf{-30.81\%} & \textbf{-41.97\%} \\
\bottomrule[1pt]
\end{tabular}
\label{table:bit_saving}
\end{table}
Then, we quantify the R-D performance by evaluating the average BD-rate~\cite{bjontegaard2001calculation} saving, which is calculated over the LPIPS results. Here, we use VVC as the anchor for comparison, where the results are shown in Table. \ref{table:bit_saving}. 
As we can see, Bai~\emph{et al.}~\cite{bai2022towards} and Chang~\emph{et al.}~\cite{chang2022conceptual} are less effective, which even require much more BD-rate than VVC (65.94\% and 162.96\% respectively). 
We can see that our MCM achieves 30.81\% on COCO and 41.97\% BD-rate savings on CelebAMask-HQ respectively, while that of the second-best approach Qian~\emph{et al.}~\cite{Yichen_2022_ICLR} is only 20.16\% on COCO and 38.68\% on CelebAMask-HQ. In general, our MCM achieves competitive bitrate cost while preserving superior reconstruction quality against existing SOTA image codecs. 

\textbf{User Study.}~
We conduct a user study and investigate the human opinion of compressed images on all of the above methods. More specifically, we randomly select 100 images from the COCO and CelebAMask-HQ validation datasets respectively. We evaluate all the methods on two low bitrates, where 50 images are included at each bitrate. 
The reconstructed images under the same dataset and the same bitrate are assigned in a group. We invite 10 volunteers to pick up the best reconstruction according to the visual qualities from the group whose images are displayed in disorder. All user study is done under the same PC to avoid visual bias. As illustrated in Fig. \ref{fig:user_study}, our MCM received the most votes on all datasets. There are more than 50\% volunteers that appreciate our results at both bitrates on COCO, whereas BPG, VVC, Qian~\emph{et al.}~\cite{Yichen_2022_ICLR} and Xie~\emph{et al.}~\cite{xie2021enhanced} get 0.8\%, 38.4\%, 3.6\%, 5.6\% votes at 0.05bpp and 1.2\%, 5.2\%, 2.3\%, 2.0\% votes at 0.08bpp on COCO. Especially, our MCM gets a definite advantage on CelebAMask-HQ, obviously demonstrating its effectiveness. 
\begin{figure}
  \centering
  \includegraphics[width=1.0\linewidth]{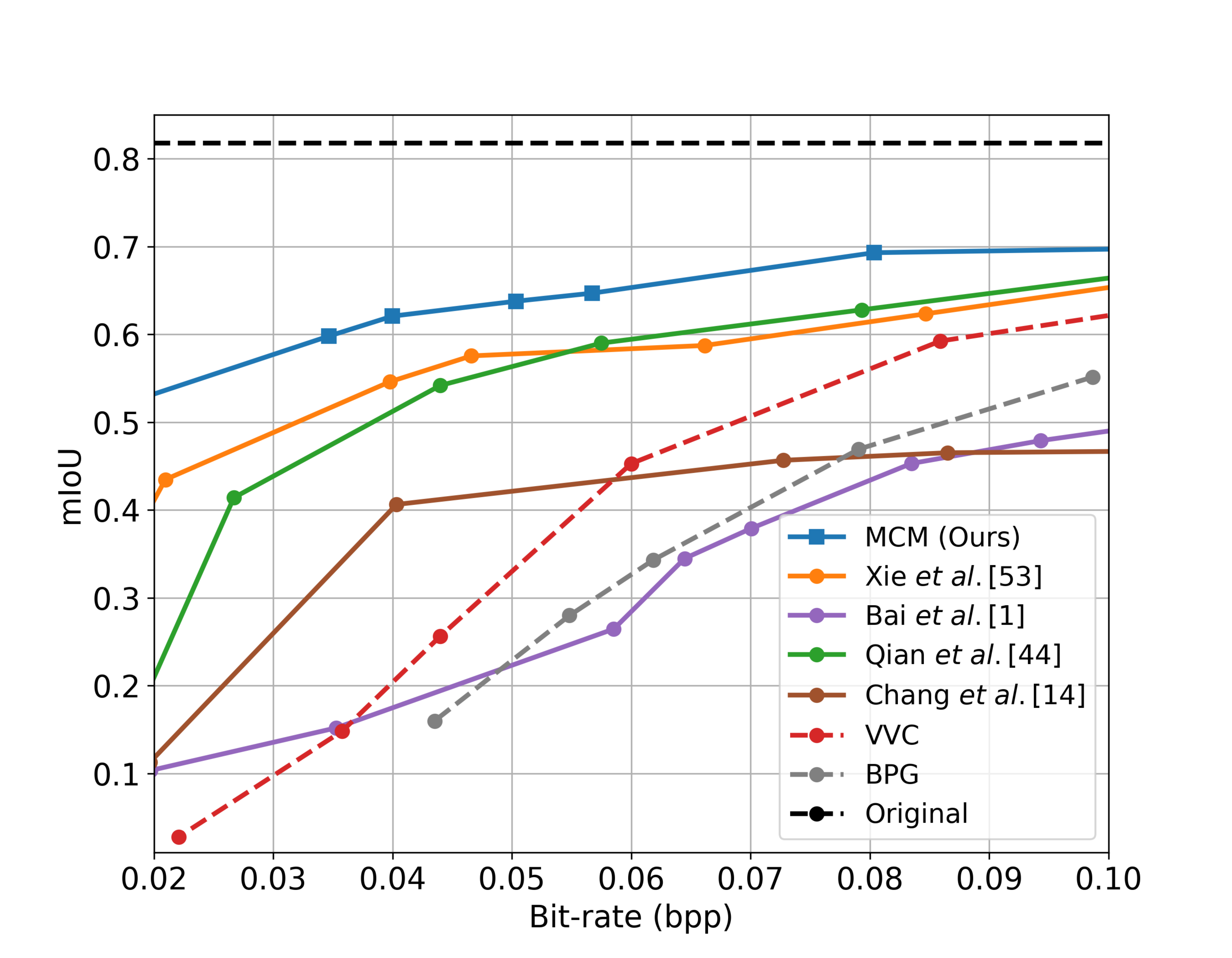}
  \vspace{-0.3cm}
  \caption{Performance of face parsing on CelebAMask-HQ testset.}
  \vspace{-0.35cm}
    \label{fig:face_machine}
\end{figure}
\subsection{Downstream Applications.}
\begin{figure*}
\hfill
\begin{overpic}[tics=20,width=\textwidth]{figures/Qualitative/0.pdf}
\put(21,4){Reference}
\put(110,4){VVC}
\put(178,4){Xie ${et}$ ${al.}$ \cite{xie2021enhanced}}
\put(259,4){Qian ${et}$ ${al.}$ \cite{Yichen_2022_ICLR}}
\put(345,4){MCM (Ours)}
\put(427,4){Ground Truth}
\end{overpic}\hfill
\begin{overpic}[width=\textwidth]{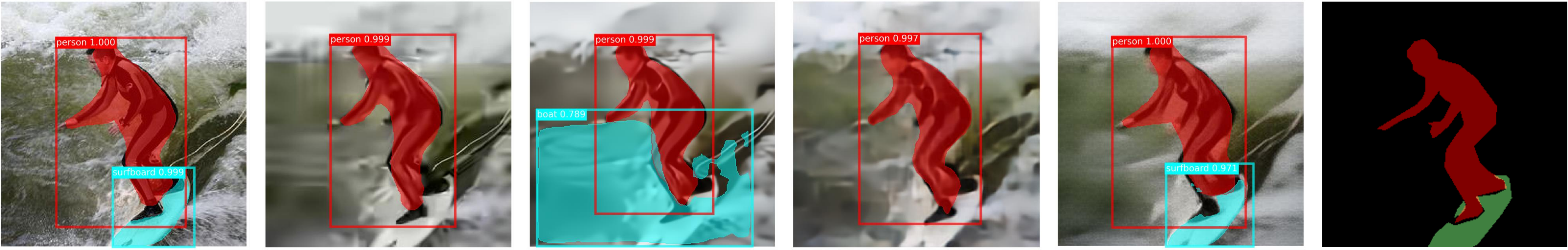}
\put(20,20){}
\end{overpic}\hfill
\begin{overpic}[tics=25,width=\textwidth]{figures/Qualitative/0.pdf}
\put(111,5){0.066}
\put(196,5){0.072}
\put(279,5){0.072}
\put(361,5){0.070}
\end{overpic}\hfill
\begin{overpic}[width=\textwidth]{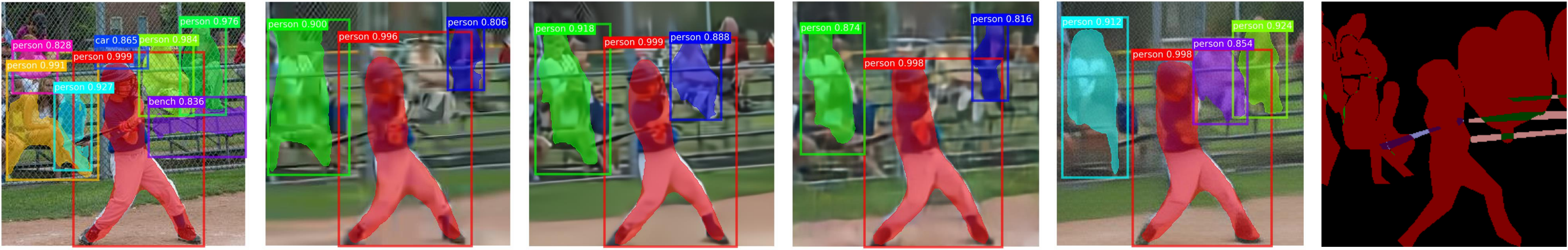}
\put(20,20){}
\end{overpic}\hfill
\begin{overpic}[tics=25,width=\textwidth]{figures/Qualitative/0.pdf}
\put(111,5){0.100}
\put(196,5){0.117}
\put(279,5){0.093}
\put(361,5){0.092}
\end{overpic}
\begin{overpic}[width=\textwidth]{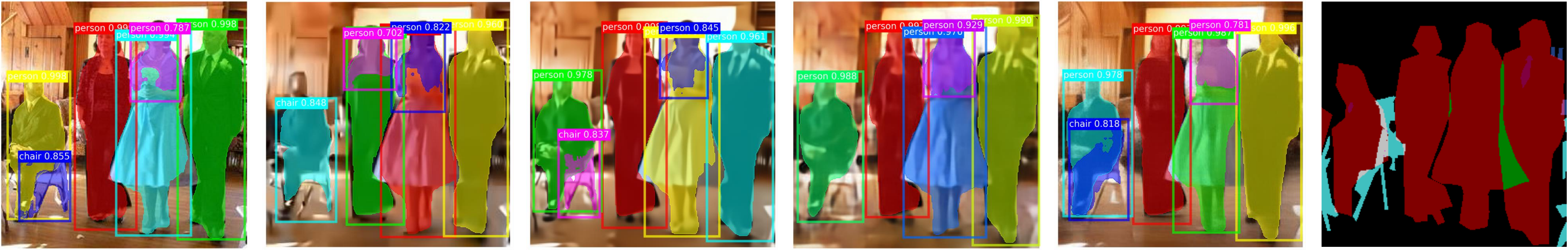}
\put(10,10){}
\end{overpic}\hfill
\begin{overpic}[tics=25,width=\textwidth]{figures/Qualitative/0.pdf}
\put(111,5){0.113}
\put(196,5){0.115}
\put(279,5){0.094}
\put(361,5){0.103}
\end{overpic}\hfill
\caption{Visual performance of instance segmentation on COCO dataset. The bitrate (bpp) of each method is also shown for comparison.}
\label{fig:segment}
\end{figure*}
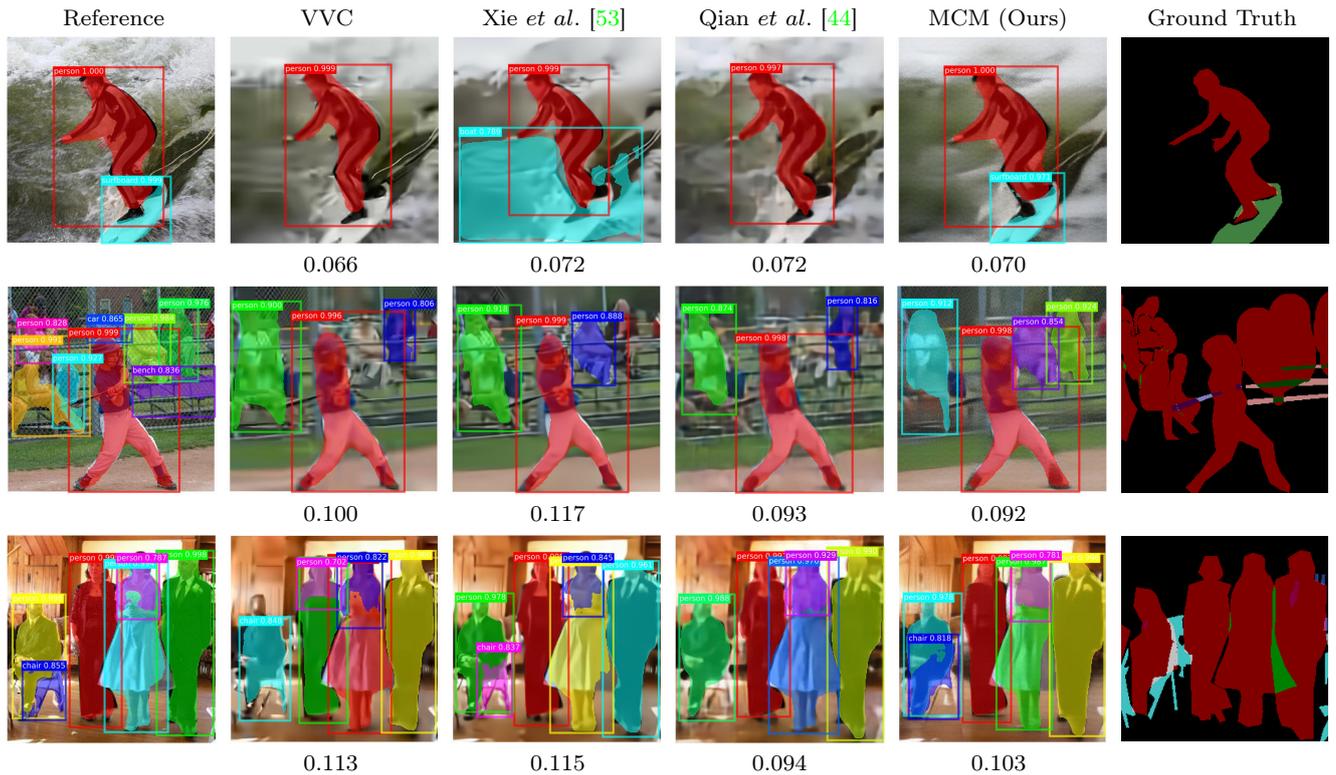
We further explore the effect of our MCM in downstream tasks using well-trained models on CelebAMask-HQ and utilize a popular face parsing method EHANet~\cite{luo2020ehanet} as the backbone to conduct compression-based face parsing. Here, we apply mIoU (mean Intersection over Union) as the metric, where the original performance of EHANet on uncompressed images is adopted as a reference. As shown in Fig. \ref{fig:face_machine}, our MCM shows excellent performance under extremely low-bitrate compression conditions: about 10\% mIoU drop down to 0.08bpp, and less than 15\% mIoU drop down to 0.055bpp. Meanwhile, all compared methods lose over 15\% mIoU already at 0.08bpp, while at 0.055bpp, they have lost about 20\%-60\% mIoU. 

Moreover, in Fig. \ref{fig:segment}, we also visualize the instance segmentation results on COCO dataset to validate the impact of MCM. We adopt pre-trained Mask R-CNN~\cite{he2017mask} as the segmentor and feed the compressed images into it to obtain the segmentation results. Besides the ground truth segmentation map, the results on original uncompressed images are used for reference. As shown in Fig. \ref{fig:segment}, in the simple example (1st row), our MCM successfully segments all the objects with better boundaries and the fewest bitrate cost. Even on images with more complex scenes and objects (last two rows), MCM still achieves more accurate segmentation performance while maintaining competitive efficiency. More surprisingly, the reference results show mislabelling of a bench in the 2nd example and overlapped persons in the 3rd example. 

\subsection{Ablation Study.}
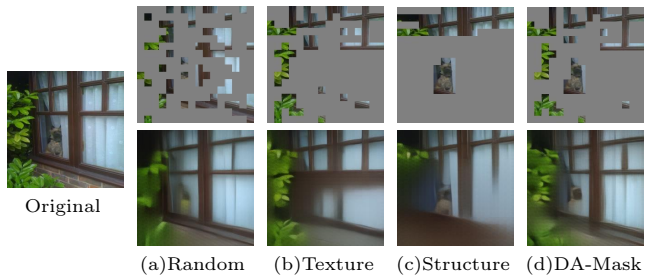
\begin{figure}[!t]
  \centering
  \begin{overpic}[tics=20,width=\linewidth]{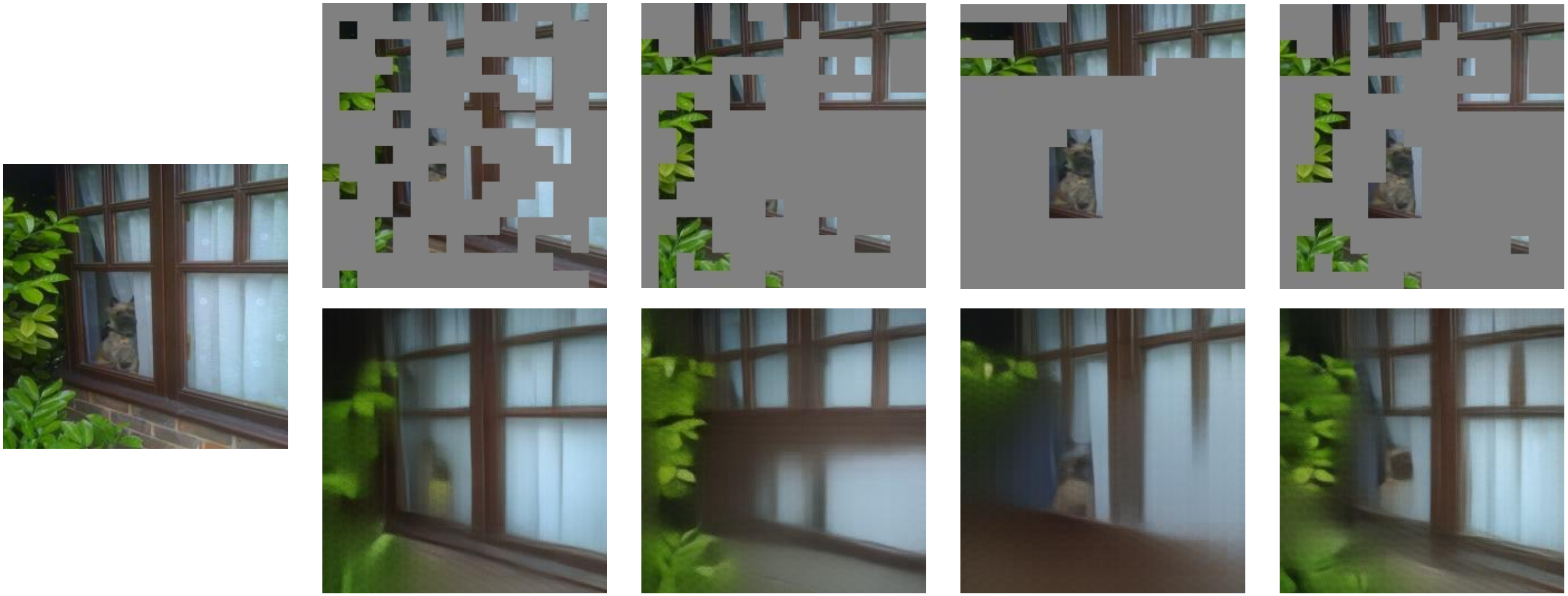}
  \put(7,13){\scriptsize{Original}}
  \end{overpic}\hfill
    \begin{overpic}[width=\linewidth]{figures/Qualitative/0.pdf}
    \put(50,1){\scriptsize{(a)Random}}
    \put(100,1){\scriptsize{(b)Texture}}
    \put(146,1){\scriptsize{(c)Structure}}
    \put(194.5,1){\scriptsize{(d)DA-Mask}}
    \end{overpic}\hfill
  \caption{Visual comparison of four mask sampling strategies. }
    \label{fig:sample_strategy}
\end{figure}

\textbf{Effect of DA-Mask.}~
\begin{table*}[t]
\caption{Quantitative comparison on COCO and CelebAMask-HQ datasets between the models with different masking strategies.}
\centering
\begin{tabular}{clcc}
\toprule[1pt]
\multicolumn{1}{c}{\textbf{Dataset}} & 
\multicolumn{1}{l}{\textbf{Sampling Strategy}} & 
\multicolumn{1}{c}{\textbf{Average bitrate↓}} & 
\multicolumn{1}{c}{\textbf{LPIPS↓}} \\ \hline
\multirow{4}{*}{COCO}  & Random  & 0.0723 & 0.263 \\
& Texture  & 0.0760 & 0.230 \\
& Structure  & 0.0737 & 0.257 \\
& DA-Mask & \textbf{0.0754} &  \textbf{0.238} \\ \hline
\multirow{4}{*}{CelebAMask-HQ} 
& Random  & 0.0839 & 0.147 \\
& Texture & 0.0885 & 0.117 \\
& Structure  & 0.0776 & 0.199 \\
& DA-Mask & \textbf{0.0803} & \textbf{0.130} \\ 
\bottomrule[1pt]
\end{tabular}
\label{table:sample_strategy}
\vspace{-0.1cm}
\end{table*}
In order to explore the effectiveness of our proposed DA-Mask strategy, we first construct three variants: 1) Random masking; 2) Texture-guided masking only, and 3) Structure-guided masking only, to make a comparison. It should be noted that all the variants are compared under the same masking ratio. 
As shown in Table \ref{table:sample_strategy}, the model with random masking involves the lowest bitrate but the worst LPIPS performance. When we replace the random masking with texture-guided strategy, the model performs slightly better than that with structure-guided masking but costs more bitrates. This is because the former focuses on the textures over the entire image while the latter only focuses on the semantic structure of the visual object. This is more pronounced on CelebAMask-HQ, in which the structural semantics in facial images are mainly described by key parts and outlines of the face, such as eyes and mouth, \emph{etc.}, thus enabling lower bitrates than other methods. 
Our DA-Mask samples the patches based on the texture and structure complexity to balance the local texture and structural semantics, which achieves a good trade-off between bitrate consumption and perceptual quality. 
To more intuitively validate the influence of different masking strategies for image compression, we illustrate their corresponding visual results in Fig. \ref{fig:sample_strategy}. As we can see, the model with random masking shows a good reconstruction effect but fails to recover the local details (Fig. \ref{fig:sample_strategy} (a)). Though using only texture-guided masking can make an improvement, it tends to ignore some semantic objects (Fig. \ref{fig:sample_strategy} (b)). In contrast, the structure-guided approach is able to retain semantic-related patches. However, it discards too many textures (Fig. \ref{fig:sample_strategy} (c)). By comparison, our DA-Mask can facilitate the model to simultaneously pay attention to important texture and semantic information of images (Fig. \ref{fig:sample_strategy} (d)), which produces more plausible images. 

\begin{figure}[t]
  \centering
    \begin{overpic}[tics=25,width=\linewidth]{figures/Qualitative/0.pdf}
    \put(51,3){75\%}
    \put(90,3){75\%}
    \put(130.5,3){43\%}
    \put(171,3){43\%}
    \put(213,3){43\%}
    \end{overpic}\hfill
  \includegraphics[width=\linewidth]{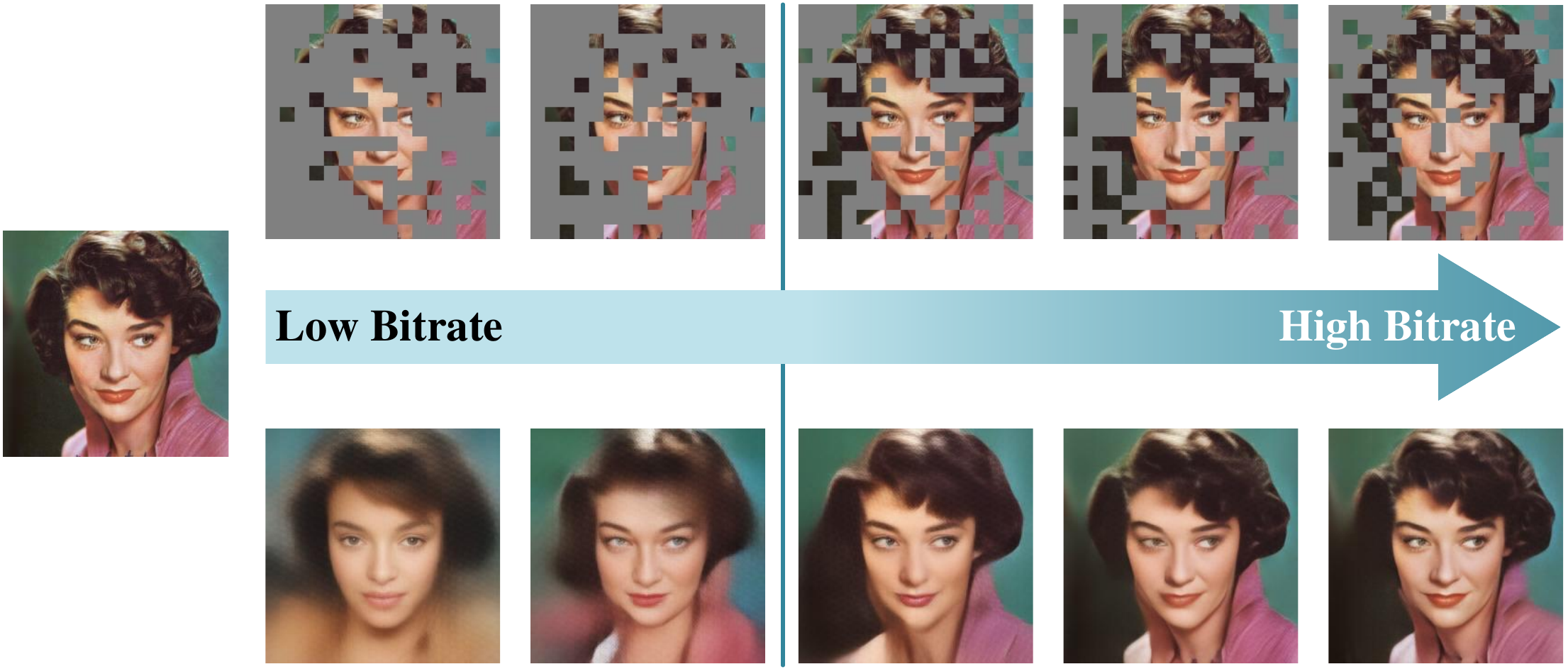}
  \begin{overpic}[tics=25,width=\linewidth]{figures/Qualitative/0.pdf}
    \put(42.5,2){$\lambda$=0.04}
    \put(81.5,2){$\lambda$=0.125}
    \put(124.5,2){$\lambda$=0.10}
    \put(165,2){$\lambda$=0.25}
    \put(211.5,2){$\lambda$=1}
    \end{overpic}
    \begin{overpic}[tics=25,width=\linewidth]{figures/Qualitative/0.pdf}
    \put(41,1){0.018bpp}
    \put(80,1){0.032bpp}
    \put(120.5,1){0.054bpp}
    \put(162,1){0.091bpp}
    \put(203.5,1){0.133bpp}
    \end{overpic}
  \caption{Impact of different masking ratio settings for image compression. From left to right, we set such values as 43\% and 75\% to select visible patches respectively. At the same masking ratio, we control the $\lambda$ in R-D loss.}
    \label{mask_more}
\end{figure}
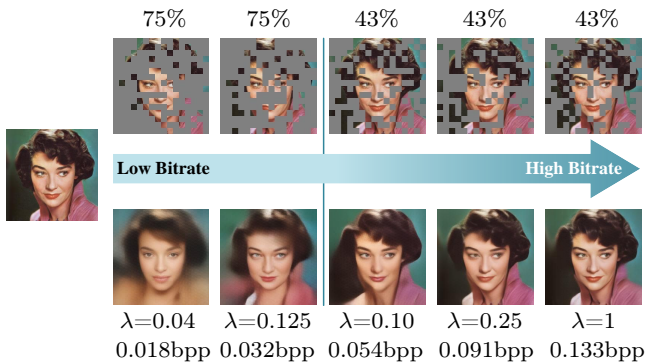

\textbf{Study of Masking ratio.}~
In this subsection, we investigate the impact of the masking ratio setting in MCM. Given a $256 \times 256$ image, we divide it into 256 patches of the size $16 \times 16$. 
The number of visible patches is set to $144$ and $64$, whose corresponding masking ratios are $43\%$ and $75\%$, respectively. 
Fig. \ref{mask_more} shows the masked input and reconstructed output with the above two masking ratios, in which the former two images are 75\% masked and the latter three images are 43\% masked. Besides, at the same masking ratio, we control the Lagrange multiplier $\lambda$ in R-D loss (shown in Eq. (\ref{equ:RD_loss})) to adjust the bitrates. Obviously, the masking ratio plays a deterministic role in both bitrate and image quality. We can find the higher ratio the more information losses, resulting in lower bitrates and poorer reconstruction quality (2nd column \emph{v.s.} 3rd column). Besides, by combining DA-Mask and the LIC module, we can jointly optimize the overall network and continuously increase the $\lambda$ values to realize quality enhancement at higher bitrates, ultimately achieving effective reconstruction while still maintaining extremely low bitrate compression (last column).
\section{Conclusion}
In this work, we have proposed a novel masked compression model (MCM) which combines the merits of pre-trained masked autoencoder (MAE) and learned image compression (LIC) for extremely low-bitrate image compression. Specifically, we present a dual-adaptive masking (DA-Mask) strategy which performs probability patch sampling based on image structure and texture distributions. This approach enables the model to adaptively sample informative patches that are conducive to image reconstruction while achieving effective redundancy removal. We note that DA-Mask with pre-trained MAE can well cooperate with LIC to learn meaningful semantic context and texture representations for effective and efficient image compression. Experimental results show that at extremely low bitrates, our MCM is superior to existing classical standard codecs and LIC methods in terms of R-D performance, human perception, and bitrate savings, and also shows better transferability to downstream tasks.

\bibliographystyle{spmpsci}      
\bibliography{main}   
\end{sloppypar}
\end{document}